\documentclass{article}

\usepackage[final, nonatbib]{neurips_2024}

\usepackage[numbers]{natbib}

\usepackage[utf8]{inputenc} 
\usepackage[T1]{fontenc}    
\usepackage{hyperref}       
\usepackage{url}            
\usepackage{booktabs}       
\usepackage{amsfonts}       
\usepackage{nicefrac}       
\usepackage{microtype}      
\usepackage{xcolor}         
\usepackage{algorithm}
\usepackage[noend]{algpseudocode}
\usepackage{enumitem}
\usepackage{sidecap}

\usepackage{microtype}
\usepackage{graphicx}
\usepackage{subfigure}
\usepackage{booktabs} 

\usepackage{amsmath}
\usepackage{amssymb}
\usepackage{mathtools}
\usepackage{amsthm}

\newcommand{\calA}{\ensuremath{\mathcal{A}}}
\newcommand{\calD}{\ensuremath{\mathcal{D}}}
\newcommand{\calK}{\ensuremath{\mathcal{K}}}
\newcommand{\calN}{\ensuremath{\mathcal{N}}}

\newcommand{\calR}{\ensuremath{\mathcal{R}}}
\newcommand{\calS}{\ensuremath{\mathcal{S}}}

\newcommand{\calL}{\ensuremath{\mathcal{L}}}

\newcommand{\Hess}{\ensuremath{\mathbf H}}

\newcommand{\GP}{{\mathcal GP}}

\newcommand{\R}{\mathbb{R}}

\newcommand{\x}{{\mathbf x}}
\newcommand{\X}{{\mathbf X}}
\newcommand{\y}{{\mathbf y}}

\newcommand{\K}{{\mathbf K}}
\newcommand{\I}{{\mathbf I}}

\newcommand{\bs}{{\mathbf s}}

\newcommand{\bu}{{\mathbf u}}

\newcommand{\D}{{\mathbf D}}
\newcommand{\w}{{\mathbf w}}

\newcommand{\btheta}{{\boldsymbol \theta}}

\newcommand{\balpha}{{\boldsymbol \alpha}}
\newcommand{\brho}{{\boldsymbol \rho}}
\newcommand{\bSigma}{{\boldsymbol \Sigma}}
\newcommand{\support}{{\mathcal{S}}}

\newcommand{\E}{{\mathbb E}}
\newcommand{\V}{{\mathbb V}}
\newcommand{\Xbb}{{\mathbb X}}

\DeclareMathOperator*{\tr}{tr}
\DeclareMathOperator*{\diag}{diag}

\newcommand{\ouralgo}{\textsc{RRP}}

\usepackage{braket}

\usepackage{amsmath}
\usepackage{amssymb}
\usepackage{mathtools}
\usepackage{amsthm}
\usepackage{thmtools, thm-restate}
\usepackage{wrapfig}
\newtheorem{thm}{Theorem}

\newtheorem{defi}[thm]{Definition}

\usepackage{hyperref}

\usepackage[textsize=tiny]{todonotes}

\title{Robust Gaussian Processes via Relevance Pursuit}

\author{%
  Sebastian Ament
  \\
  Meta\\
  \texttt{ament@meta.com} \\
  \And
  Elizabeth Santorella \\
  Meta \\
  \texttt{santorella@meta.com} \\
  \And
  David Eriksson \\
  Meta \\
  \texttt{deriksson@meta.com}
  \And
  Ben Letham \\
  Meta \\
  \texttt{bletham@meta.com}
  \And
  Maximilian Balandat \\
  Meta \\
  \texttt{balandat@meta.com} \\
  \And
  Eytan Bakshy \\
  Meta \\
  \texttt{ebakshy@meta.com}
}

\begin{document}

\maketitle


\begin{abstract}
  Gaussian processes (GPs) are non-parametric probabilistic regression models that are popular due to their flexibility, data efficiency, and well-calibrated uncertainty estimates. However, standard GP models assume homoskedastic Gaussian noise, while many real-world applications are subject to non-Gaussian corruptions. Variants of GPs that are more robust to alternative noise models have been proposed, and entail significant trade-offs between accuracy and robustness, and between computational requirements and theoretical guarantees. In this work, we propose and study a GP model that achieves robustness against sparse outliers by inferring data-point-specific noise levels with a sequential selection procedure maximizing the log marginal likelihood that we refer to as \emph{relevance pursuit}. 
  We show, surprisingly, that the model can be parameterized such that the associated log marginal likelihood is {\it strongly concave} in the data-point-specific noise variances, a property rarely found in either robust regression objectives or GP marginal likelihoods. This in turn implies the weak submodularity of the corresponding subset selection problem, and thereby proves approximation guarantees for the proposed algorithm. 
  We compare the model's performance relative to other approaches on diverse regression and Bayesian optimization tasks, including the challenging but common setting of sparse corruptions of the labels within or close to the function range.
\end{abstract}

\section{Introduction}
\label{sec:Introduction}

Probabilistic models have long been a central part of machine learning, 
and 
Gaussian process (GP) models are a key workhorse for many important tasks \citep{rasmussen2006gaussian}, especially in the small-data regime. 
GPs are flexible, non-parametric predictive models known for their high data efficiency and well-calibrated uncertainty estimates, making them a popular choice for regression, uncertainty quantification, and downstream applications such as Bayesian optimization~(BO) \citep{ament2023sustainableconcretebayesianoptimization, frazier2018tutorial, garnett_bayesoptbook_2023} and active learning \citep{ament2021sara, riis2022bayesian}.

GPs flexibly model a distribution over functions, but assume a particular observation model.
The standard formulation assumes i.i.d Gaussian observation noise, i.e., $y(\x) = f(\x) + \epsilon$, where $f(\x)$ is the true (latent) function value at a point $\x$ and $\epsilon \sim \mathcal{N}(0, \sigma^2)$, implying a homoskedastic Gaussian likelihood. 
While mathematically convenient, this assumption can be a limitation in practice, since noise distributions are often heavy-tailed or observations may be corrupted due to issues such as sensor failures, data processing errors, or software bugs. 
Using a standard GP model in such settings can result in poor predictive performance.

A number of \textit{robust} GP modeling approaches have been proposed to remedy this shortcoming, most of which fall into the following broad categories: data pre-processing (e.g., Winsorizing), modified likelihood functions (e.g., Student-$t$), and model-based data selection and down-weighting procedures. 
These approaches offer different trade-offs between model accuracy, degree of robustness, broad applicability, computational requirements, and theoretical guarantees.

\begin{wrapfigure}{r}{0.5\textwidth}
    \centering
    \vspace{-20pt}
    \includegraphics[width=0.5\textwidth]{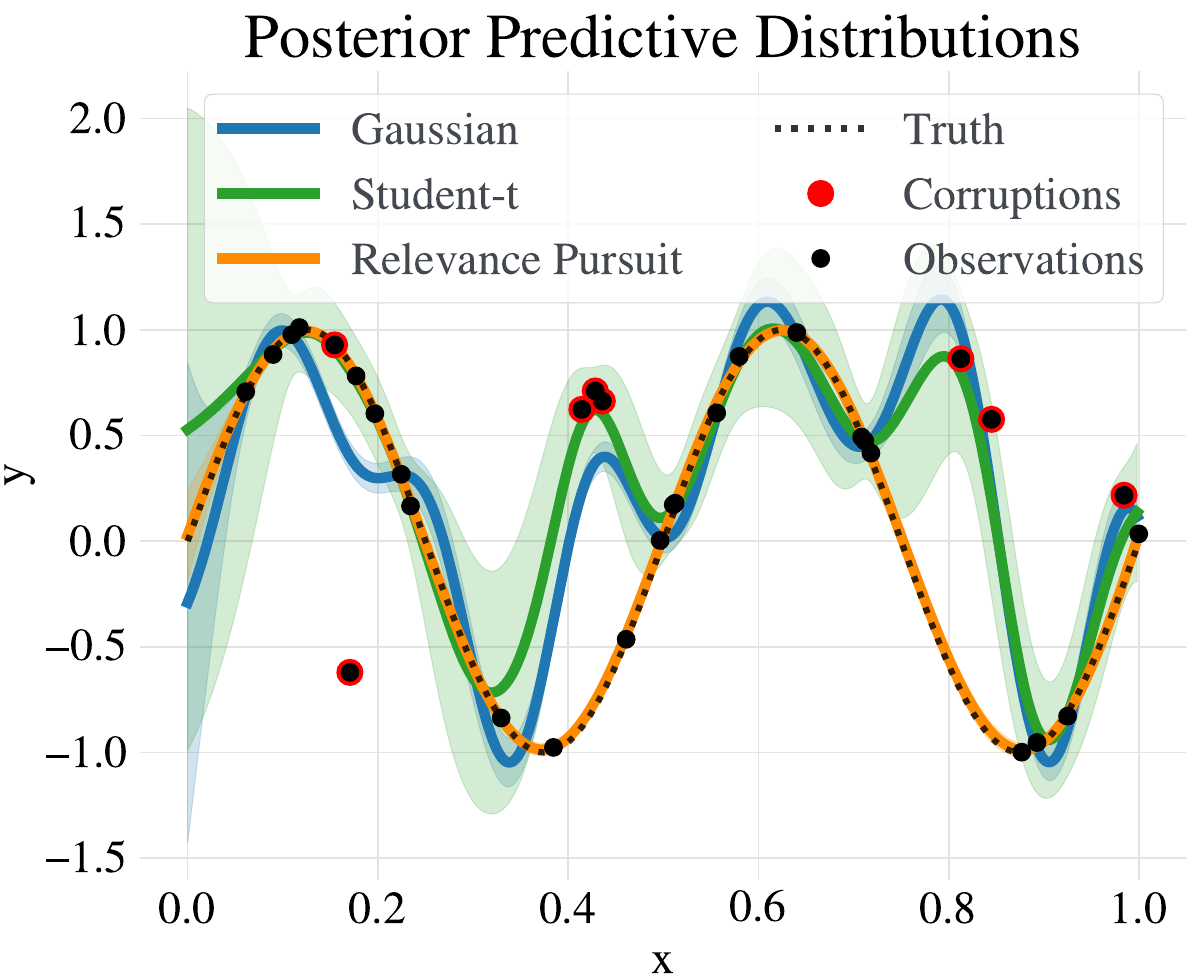}
    \vspace{-20pt}
    \caption{Comparison of \ouralgo{} to a standard GP and a variational GP with a Student-$t$ likelihood on a regression example. While the other models are led astray by the corrupted observations, \ouralgo{} successfully identifies the corruptions (red) and thus achieves a much better fit to the ground truth.}
    \vspace{-20pt}
    \label{fig:sine_example}
\end{wrapfigure}

In this paper, we propose a simple yet effective implicit data-weighting approach that endows GPs with a high degree of robustness to challenging label corruptions. 
Our approach is flexible and can be used with arbitrary kernels, is efficient to compute, and yields provable approximation guarantees.
Our main contributions are as follows:
\begin{enumerate}[leftmargin=18pt, labelwidth=!, labelindent=0pt, itemsep=2pt]
    \item We propose a modification to the standard GP model that introduces learnable data-point-specific noise variances.
    \item We introduce a novel greedy sequential selection procedure for maximizing the model's marginal log-likelihood (MLL) that we refer to as \emph{relevance pursuit}.
    \item We prove that, under a particular parameterization, the MLL is strongly concave in the data-point-specific noise variances, and derive approximation guarantees for our algorithm.
    \item We demonstrate that our approach, Robust Gaussian Processes via Relevance Pursuit (\ouralgo{}), performs favorably compared to alternative methods across various benchmarks, including challenging settings of sparse label corruptions within the function's range,
    see e.g. Figure~\ref{fig:sine_example}.
\end{enumerate}


\section{Preliminaries}
\label{sec:Preliminaries}

We aim to model a function $f: \mathbb{X} \rightarrow \R$ over some domain $\mathbb{X} \subset \R^d$. 
With a standard Gaussian noise model, for $\x_i \in \mathbb{X}$ we obtain observations $y_i = f(\x_i) + \epsilon_i$, where $\epsilon_i \sim \calN(0, \sigma^2)$ are i.i.d. draws from a Gaussian random variable. 
$\|\cdot\|$ denotes the Euclidean norm unless indicated otherwise.

\subsection{Gaussian Processes}
\label{subsec:Preliminaries:GPs}
A GP $f \sim \GP(\mu(\cdot), k_{\btheta}(\cdot,\cdot))$ is fully defined by its mean function $\mu: \Xbb \rightarrow \R$ and covariance or kernel function $k_{\btheta}: \Xbb \times \Xbb \rightarrow \R$, which is parameterized by $\btheta$. 
Without loss of generality, we will assume that $\mu \equiv 0$. 
Suppose we have collected data $\calD = \{(\x_i, y_i)\}_{i=1}^n$ where $\X := \{\x_i\}_{i=1}^n$, $\y := \{y_i\}_{i=1}^n$. 
Let $\bSigma_\btheta \in \calS_{++}^n$ denote the covariance matrix of the data set, i.e., $[\bSigma_\btheta]_{ij} = k_\btheta(\x_i, \x_j) + \delta_{ij} \sigma^2$, where $\delta_{ij}$ is the Kronecker delta.
The negative marginal log-likelihood (NMLL) $\calL$ is given by
\begin{align}
    \label{eq:Preliminaries:GPs:mll}
    -2\calL(\btheta) :=
    -2 \log p(\y| \X, \btheta) = \y^\top \bSigma_\btheta^{-1} \y 
        + \log \det \bSigma_\btheta + n \log 2 \pi.
\end{align}

In the following, we will suppress the explicit dependence of the kernel matrix on $\btheta$ for brevity of notation. 
For a comprehensive background on GPs, we refer to \citet{rasmussen2006gaussian}.

\subsection{Noise Models}
\label{subsec:Preliminaries:NoiseModels}

\paragraph{Additive, heavy-tailed noise}
Instead of assuming the noise term $\epsilon_i$ in the observation model to be Gaussian, other noise models consider zero-mean perturbations drawn from distributions with heavier tails, such as the Student-$t$~\citep{jylanki2011robust}, Laplace~\citep{kuss2006gaussian},
or $\alpha$-Stable~\citep{ament2018accurate} distributions. 
These types of errors are common in applications such as finance, geophysics, and epidemiology~\citep{congdon2017representing}.  
Robust regression models utilizing Student-$t$ errors are commonly used to combat heavy-tailed noise and outliers.

\paragraph{Sparse corruptions}
In practice, often a small number of labels are corrupted. 
We will refer to these as ``outliers," though emphasize that the corrupted values may fall within the range of normal outputs. 
Sparse corruptions are captured by a model of the form $y_i = Z_i f(\x_i) + (1 - Z_i)W_i$, where $Z_i \in \{0, 1\}$ and $W_i \in \R$ is a random variable. 
Note that $W_i$ need not have (and rarely has) $f(\x_i)$ as its mean. 
For instance, consider a faulty sensor that with some probability $p$ reports a random value within the sensor range $[y_l, y_h]$. In this case $Z_i \sim \text{Ber}(p)$ and $W_i \sim \text{U}[y_l, y_h]$.
Software bugs, such as those found in ML training procedures, or errors in logging data can result in sparse corruptions.

\section{Related Work}
\label{sec:RelatedWork}

\paragraph{Data pre-processing}
Data pre-processing can be an effective technique for handling simple forms of data corruption, such as values that fall outside a valid range of outputs. With such pre-processing, outliers are handled upstream of the regression model. 
Common techniques include the power transformations~\citep{box1964analysis}, trimming, and winsorization. 
These methods can add substantial bias if not used carefully, and generally do not handle data corruptions that occur within the normal range of the process to be modeled. 
See~\citep{chu2016cleaning} for a review on data cleaning.


\paragraph{Heavy-tailed likelihoods} 
One class of robust methods uses additive heavy-tailed noise likelihoods for GPs, particularly Student-$t$~\citep{jylanki2011robust}, Laplace~\citep{ranjan2016robustGPEMalgo},
and Huber~\citep{algikar2023robustgaussianprocessregression},
and could be extended with $\alpha$-Stable distributions, 
which follow a generalized central limit theorem~\citep{ament2018accurate}.
These models are less sensitive to outliers, but they lose efficiency when the outliers are a sparse subset of the observations, as opposed to global heavy-tailed noise. Furthermore, model inference is no longer analytic, necessitating the use of approximate inference approaches such as MCMC~\citep{neal1997mcmcGP}, Laplace approximation~\citep{vanhatalo2009studenttGP}, expectation propagation (EP) \citep{jylanki2011robust}, Expectation Maximization \citep{ranjan2016robustGPEMalgo}, or variational inference~\citep{tipping2005variationalstudentt}. 
\citet{shah2014student} take a related approach using a Student-$t$ process prior in the place of the GP prior. Unfortunately, the Student-$t$ process is not closed under addition and lacks the tractability that makes GPs so versatile.
Alternative noise specifications include a hierarchical mixture of Gaussians \citep{daemi2019gpmixturelik} and a ``twinned'' GP model \citep{Naish2007twinnedgp} that uses a two-component noise model to allow outlier behavior to depend on the inputs.
This method is suited for settings where outliers are not totally stochastic, but generally is not able to differentiate ``inliers'' from outliers when they can occur with similar inputs.


\paragraph{Outlier classification} 

\citet{awasthi2022trimmed} introduces the Trimmed MLE approach, which identifies the subset of data points (of pre-specified size) under which the marginal likelihood is maximized. 
\citet{andrade2023trimmed} fit GPs using the trimmed MLE by applying a projected gradient method to an approximation of the marginal likelihood. The associated theory only guarantees convergence to a stationary point, with no guarantee on quality. When no outliers are present, this method can be worse than a standard GP.   
%
\citet{li2021iterativetrimming} propose a heuristic iterative procedure of removing those data points with the largest residuals after fitting a standard GP, with subsequent reweighting. The method shows favorable empirical performance but has no theoretical guarantees, and fails if the largest residual is not associated with an outlier.
\citet{park2022robustGPRbias} consider a model of the form $y_i = \delta_i + f(\x_i) + \epsilon_i$, where outliers are regarded as data with a large bias $\delta_i$. 
Their random bias model is related to our model in that it also introduces learnable, data-point-specific variances. However, inference is done in one step by optimizing the NMLL with an inverse-gamma prior on the $\delta_i$'s,
which -- in contrast to the method proposed herein -- generally does not lead to exactly sparse $\delta_i$'s .

\paragraph{Sample re-weighting}
\citet{altamirano2023robust} propose robust and conjugate GPs (RCGP) based on a modification to the Gaussian likelihood function that is equivalent to standard GP inference,
where the covariance of the noise $\sigma^2 \I$ is replaced by $\sigma^2 \diag(\w^{-2})$ and the prior mean $\mathbf m$ is replaced by $\mathbf m_\w = \mathbf m + \sigma^2 \nabla_y \log(\w^2)$. 
The authors advocate for the use of the inverse multi-quadratic weight function $w(\x, y) = \beta (1 + (y - m(\x))^2 / c^2)^{-1/2}$, which introduces two additional hyper-parameters: the soft threshold $c$, and the ``learning rate'' $\beta$. 
Importantly, the weights $\w$ are defined {\it a-priori} as a function of the prior mean $m(\x)$ and the targets $y$, thereby necessitating the weights to identify the correct outliers
without access to a model. This is generally only realistic if the outlier data points are clearly separated in the input or output spaces 
rather than randomly interspersed. 




\section{Robust Gaussian Process Regression via Relevance Pursuit}
\label{sec:rgp-rp}

Our method adaptively identifies a sparse set of outlying data points that are corrupted by a mechanism that is not captured by the other components of the model. This is in contrast to many other approaches to robust regression
 that non-adaptively apply a heavy-tailed likelihood to {\it all} observations, which can be suboptimal if many observations are of high quality.

\subsection{The Extended Likelihood Model}
\label{sec:Theory:RobustVariance}

We extend the standard GP observation noise variance $\sigma^2$  with data-point-specific noise variances $\brho\ = \{\rho_i\}_{i=1}^n$, so that the $i$-th data point is distributed as 
\begin{equation}
\label{eq:basic_robust_likelihood}
    y_i \ \big | \ \x_i \sim \calN
    \left(f(\x_i), \sigma^2 + \rho_i\right).
\end{equation}
This is similar to Sparse Bayesian Learning \citep{tipping2001sparse} 
in which weight-specific prior variances control a feature's degree of influence on a model's predictions.
The marginal likelihood optimization of $\rho_i$
in \eqref{eq:basic_robust_likelihood}
gives rise to an {\it automatic mechanism} for the detection and weighting of outliers.
The effect of $y_i$ on the estimate of $f$ vanishes as $\rho_i \rightarrow \infty$, similar to the effect of the latent varibales $\mathbf h$ in 
\citet{bodin2020modulatingsurrogates}'s extended GP model $f(\x, \mathbf h)$,
though $\mathbf h$ requires MCMC for inference.
While many heteroskedastic GP likelihoods model noise as an input-dependent process \citep{goldberg1998inputnoiseGP, kersting2007mostlikelyhet}, 
our formulation does not require such assumptions, 
and is thus suitable for corruptions that
are not spatially correlated. 


An elegant consequence of our modeling assumption is
that we can compute individual marginal-likelihood maximizing 
$\rho_i$'s in closed form
when keeping all $\rho_j$ for $j \neq i$ fixed. 
In particular,

\begin{restatable}{lem}{robustvariance}[Optimal Robust Variances]
\label{lem:robust_incremental_ml}
Let $\calD_{\backslash i} = \{(\x_j, y_j): j \neq i\}$, 
$\brho = \brho_{\backslash i} + \rho_i \mathbf e_i$,
where $\brho, \brho_{\backslash i} \in \R_+^n$,
$[\brho_{\backslash i}]_i = 0$,
and $\mathbf e_i$ is the $i$th canonical basis vector.
Then keeping $\brho_{\backslash i}$ fixed, 
\begin{equation}
\label{eq:robust_incremental_ml}
    \begin{aligned}
     \rho_i^* = \arg \max_{\rho_i} \calL \bigl(\brho_{\backslash i} + \rho_i \mathbf e_{i} \bigr) 
     = 
    \left[
        (y_i - \mathbb{E}[y(\x_i) | \calD_{\backslash i}])^2  
        - \mathbb{V}[y(\x_i) | \calD_{\backslash i}]
        \right]_+,
    \end{aligned}
\end{equation}
where $y(\x_i) = f(\x_i) + \epsilon_i$.
These quantities can be expressed as functions of $\bSigma^{-1} = (\K + \D_{\sigma^2 + \brho})^{-1}$:
\begin{equation}
\nonumber
    \begin{aligned}
        \E[y(\x_i) | \calD_{\backslash i}]^2 = y_i - \left[
            \bSigma^{-1} \y
            \right]_i
            \big / \left[ \bSigma^{-1} \right]_{ii}, 
            \qquad 
            \text{and}
            \qquad
        \V[y(\x_i) | \calD_{\backslash i}] = 1 \big / \left[ \bSigma^{-1} \right]_{ii},
    \end{aligned}
\end{equation} 
where $\D_{\sigma^2 + \brho}$ is a diagonal matrix whose entries are ${\sigma^2 + \brho}$. 
\end{restatable}

The first component $\mathbb{E}[f(\x_i) + \epsilon_i | \calD_{\backslash i}]^2$ of~\eqref{eq:robust_incremental_ml} is the empirical error to $y_i$ of the model trained without the $i$-th data point, i.e., the leave-one-out (LOO) cross-validation error~\citep{rasmussen2006gaussian}. 
The second component $\mathbb{V}[f(\x_i) + \epsilon_i | \calD_{\backslash i}]$ is the LOO predictive variance. 
The optimal solution to $\rho_i$ is only non-zero for those observations whose squared LOO error is larger 
than the
LOO predictive variance at that point.


%
\begin{figure}[t!]
    \centering 
    \includegraphics[width=0.99\textwidth]{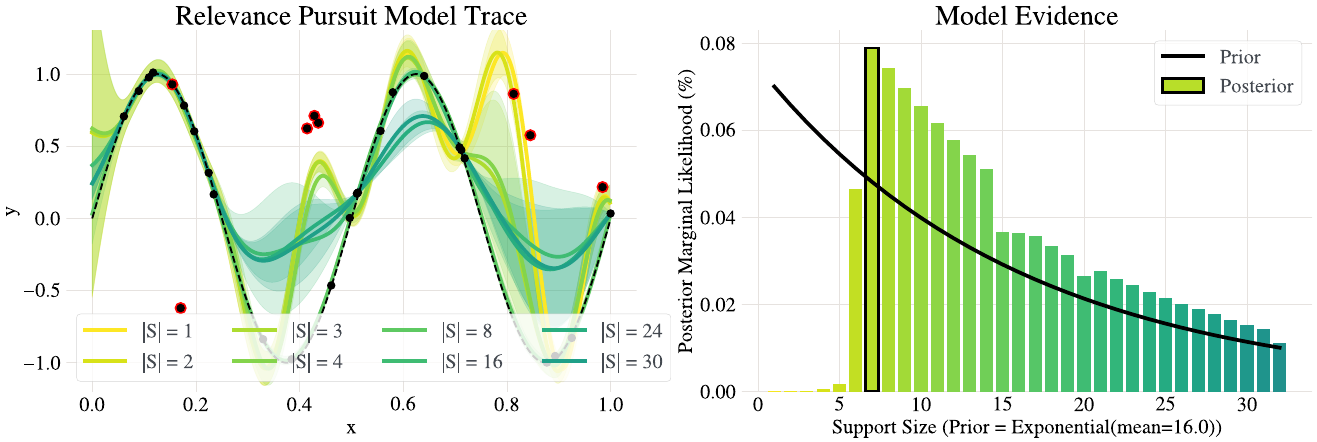}
    \caption{\textit{Left:} Evolution of model posterior during Relevance Pursuit, 
    as the number of data-point-specific variances $|S|$ increases (from light colors to dark). Red points indicate corruptions that were generated by uniformly sampling from the function's range.
    \textit{Right:} Comparison of posterior marginal likelihoods as a function of a model's $|S|$. The maximizer -- boxed in black -- is the preferred model.}
    \label{fig:enter-label}
\end{figure}

\subsection{Optimization with a Maximum Number of Outliers}
\label{sec:CappedAlgorithmDescription}

Without additional structure, inference of the noise variances $\rho_i$ 
does not yield desirable models, 
as the marginal likelihood can be improved by increasing the prior variance $\rho_i$
of any data point where 
Eq.~\eqref{eq:robust_incremental_ml} is greater than zero,
even if that is due to regular (non-outlier) measurement noise.
To avoid this, we constrain the number of non-zero $\rho_i$, that is, $\|\brho\|_\infty = |\{0 < \rho_i \}| \leq k < n$.  
While this sparsity constraint mitigates over-flexibility,
it gives rise to a formidably challenging optimization problem, as there are a combinatorial number of sparse outlier sets to consider.
Even if the number of outliers $n_o$ were known, exhaustive search would still require considering $n$-choose-$n_o$ possibilities. 

For tractability, we iteratively add data points to a set of potential ``outliers''
by setting their associated $\rho_i$ to be nonzero,
using the closed-form expression for the optimal individual $\rho_i$ variances
in Lemma~\ref{lem:robust_incremental_ml}.
As the algorithm seeks to identify the most ``relevant'' data points (as measured by $\calL$) upon completion, we refer to it as \emph{Relevance Pursuit}. This is Algorithm~\ref{algo:RobustRegressionModel:Inference:RelevancePursuitForward} with $\mathrm{useBayesianModelSelection}$ as $\mathrm{false}$.
Specifically, this is the ``forward'' variant; Algorithm~\ref{algo:RobustRegressionModel:Inference:RelevancePursuitBackward} in the Appendix presents an alternative ``backward'' variant that we found to work well if the number of corrupted data points is large. 

Crucial to the performance of the optimizer,
it never removes data from consideration completely; a data point is only down-weighted if it is apparently an outlier. This allows the down-weighting to be reversed if a data point appears ``inlying'' after having down-weighted other data points, improving the method's robustness and performance. 
This is in contrast to \citet{andrade2023trimmed}'s greedy algorithm, in which the exclusion of a data point can both increase or decrease the associated marginal likelihood. 
This means that their objective is not monotonic, a necessary condition to provide constant-factor submodular approximation guarantees for greedy algorithms, see Section~\ref{sec:Theory}.\\[-2ex]

\begin{algorithm}[h!]
\caption{Relevance Pursuit (Forward Algorithm)}
\label{algo:RobustRegressionModel:Inference:RelevancePursuitForward}
\begin{algorithmic}
    \Require $\X$, $\y$, schedule $\calK =(k_1, k_2, \dotsc, k_{\calK})$, $\mathrm{useBayesianModelSelection}$ (boolean)
    \State Initialize $\support_0 \subseteq \{1, \hdots, n\}$ (typically $\support_0 = \emptyset$)
    \For{$i$ in $(1, \dotsc, |\calK|)$}
        \State Optimize MLL: $\brho_{\support_i} \leftarrow \arg \max_{\brho_{\support_i}} 
        \calL \left(\brho_{\support_i} \right)$, \; where $\brho_{\support_i} = \{\brho: \rho_j = 0, \; \forall \, j \not \in \support_i \}$.
        \State Expand Support: 
        \State \qquad Compute $\Delta_i(j) \leftarrow \max_{\rho_j} \calL(\brho_{\support_i} + \rho_j \mathbf e_{j}) - \calL(\brho_{\support_i})$ for all $j \not \in \support_i$ via Lemma~\ref{lem:robust_incremental_ml} .
        \State \qquad $\calA_i \leftarrow \{j_1, \dotsc, j_{k_i}\}$ 
        such that $\Delta_i(j) \geq \Delta_i(j')$ for all $j \in \calA_i$
        and $j' \not \in (\calA_i \cup \support_i)$.
        \State \qquad $\support_{i + 1} \leftarrow \support_i \cup \calA_i$
    \EndFor
    \If{$\mathrm{useBayesianModelSelection}$}
        \State Compute the marginal likelihood $p(\calD |\support_i) \approx  p(\calD | \support_i, \brho_{\support_i})$
        \State $\support^* \leftarrow \arg \max_{\support_i} p(\calD | \support_i ) p(\support_i)$.
    \Else
        \State $\support^*$ = $\support_{\calK}$. 
    \EndIf
    \State Return $\support^*$, $\brho_{\support^*}$.
\end{algorithmic}
\end{algorithm}
\vspace{-1.75ex}

\subsection{Automatic Outlier Detection via Bayesian Model Selection}
\label{sec:MainAlgorithmDescription}

In practice, it is often impossible to set a hard threshold on the number of outliers for a particular data set.
For example, a sensor might have a known failure rate, but how many outliers it produces will depend on the specific application of the sensor.
Thus, is often more natural to specify a prior distribution $p(\support)$ over the number of outliers, rather than fix the number \textit{a priori}. We leverage the Bayesian model selection framework
\citep{wasserman2000bayesian, 
lotfi2023bayesianmodelselectionmarginal}
to determine the most probable
number of outliers in a data- and model-dependent way, aiming to maximize $p(\support | \calD)$.
This gives rise to Algorithm~\ref{algo:RobustRegressionModel:Inference:RelevancePursuitForward}, with $\mathrm{useBayesianModelSelction}$ as $\mathrm{true}$.

Computationally, we start by iteratively adding outliers up to the maximal support of the prior, similar to the procedure described in Section~\ref{sec:CappedAlgorithmDescription}. We store a trace of models generated at each iteration, then approximate the model posterior
$p(\support_i | \calD) \propto p(\calD | S_i) p(S_i)$ at each point in the trace. 
As the exact posterior is intractable, we approximate it with 
$p(\calD | \support_i) = \int p(\calD | \support_i, \brho_{\support_i}) \text{d}\brho_{\support_i}
\approx  p(\calD | \support_i, \brho_{\support_i}^*)$. Finally, we select the model from the model trace $\{\support_i\}_{i}$
that attains the highest model posterior likelihood. 
%
Imposing a prior on the number of outliers differs notably from most 
sparsity-inducing priors,
which are instead defined on the parameter values, like $l_1$-norm regularization.
In practice, $p(\support)$ can be informed by empirical distributions of outliers.
For our experiments, we use an exponential prior on $|\support|$ to encourage the selection of models 
that fit as much of the data as tightly as possible.




Regarding the schedule $\calK$ in Algorithm~\ref{algo:RobustRegressionModel:Inference:RelevancePursuitForward}, the most natural choice is simply to add one data point at a time, i.e. $\calK = (1, 1, ...)$, but this can be slow for large $n$. 
In practice, we recommend schedules that test a fixed set of outlier fractions, e.g. $\calK = (0.05n, 0.05n, \dots)$.


\section{Theoretical Analysis}
\label{sec:Theory}

We now provide a theoretical analysis of our approach. 
We first propose a re-parameterization of the $\rho_i$ that maps the optimization problem to a compact domain. 
Surprisingly, the re-parameterized problem exhibits strong convexity and smoothness when the base covariance matrix (excluding the~$\rho_i$) is well-conditioned. 
We connect the convexity and smoothness with existing results that yield approximation guarantees for sequential greedy algorithms,
implying a 
constant-factor approximation guarantee to the optimal achievable NMLL value for generalized orthogonal matching pursuit (OMP),
a greedy algorithm that is closely related to
Algorithm~\ref{algo:RobustRegressionModel:Inference:RelevancePursuitForward}.


\subsection{Preliminaries for Sparse Optimization}
\label{sec:Theory:SparseOpt}

The optimization of linear models with respect to least-squares objectives in the presence of sparsity constraints has been richly studied in statistics~\citep{tibshirani1996regression},
compressed sensing~\citep{ament2021optimality, tropp2004greed}, 
and machine learning~\citep{ament2021sblviastepwise, wipf2007newvieward}. 
Of central importance to the theoretical study of this problem class
are the eigenvalues of sub-matrices of the feature matrix,
corresponding to sparse feature selections and so often referred to as {\it sparse eigenvalues}.
The restricted isometry property (RIP) formalizes this.

\begin{defi}[Restricted Isometry Property]
\label{defi:rip}
An $(n \times m)$-matrix $\mathbf A$ satisfies the $r$-restricted isometry property (RIP)
with constant $\delta_r \in (0, 1)$ 
if for every submatrix $\mathbf A_\support$ with $|\support| = r \leq m$ columns,
\[
(1-\delta_r) \|\x\| \leq \|\mathbf A_\support \x_{\support} \| \leq (1 + \delta_r) \|\x\|,
\]
where $\x_\support \in \R^r$.
This is equivalent to 
$(1 - \delta_r) \I \preceq (\mathbf A_\support^* \mathbf A_\support) \preceq (1 - \delta_r) \I$.
\end{defi}

The RIP has been proven to lead to  exact recovery guarantees~\citep{candes2008restricted}, as well as approximation guarantees~\citep{das2018approximate}.
\citet{elenberg2018restricted} generalized the RIP to non-linear models and other data likelihoods, 
using the notion of restricted strong convexity (RSC) and restricted smoothness.

\begin{defi}[Restricted Strong Convexity and Smoothness]
\label{defi:rsc}
A function $f: \R^d \to \R$ is $m_r$-restricted strong convex
and $M_r$-restricted smooth if for all $(\x, \x')$
in the domain 
$D_r \subset (\R^d \times \R^d)$,
    \[
    m_r \| \x' - \x \|^2 / 2
    \ \leq \ 
    f(\x') - f(\x) - 
    \nabla[f](\x)^\top (\x' - \x)
    \ \leq \
    M_r \| \x' - \x\|^2 / 2.
    \]
In the context of sparse optimization, we let 
$D_r$ be the set of tuples of $r$-sparse 
vectors whose difference is also at most $r$-sparse.
In particular,
$D_r =
\{
    (\x, \x') \ \text{s.t.} \ 
    \|\x\|_0, 
    \|\x'\|_0, 
    \|\x' - \x\|_0 \leq r
\}
$.
\end{defi}


Generalized orthogonal matching pursuit (OMP)
\citep{ament2022gmp, locatello2017gmp, locatello2018matchingpursuit}
is a greedy algorithm
that keeps track of a support set $\support$ of non-zero coefficients,
and expands the support based on the largest gradient magnitudes,
applied to the marginal liklihood optimization problem,
$\support_{i + 1} = \support_{i} \cup \arg \max_{j \not \in \support} |\nabla_{\brho}\calL(\brho)|_j$.
Algorithm~\ref{algo:RobustRegressionModel:Inference:RelevancePursuitForward}
generalizes OMP~\citep{tropp2004greed} by allowing more general support expansion schedules~$\calK$, 
and specializes the support expansion criterion using the 
special problem structure exposed by Lemma~\ref{lem:robust_incremental_ml}.



\subsection{The Convex Parameterization}
\label{sec:Theory:Convexity}

\begin{figure}
    \centering
    \includegraphics[width=0.99\textwidth]{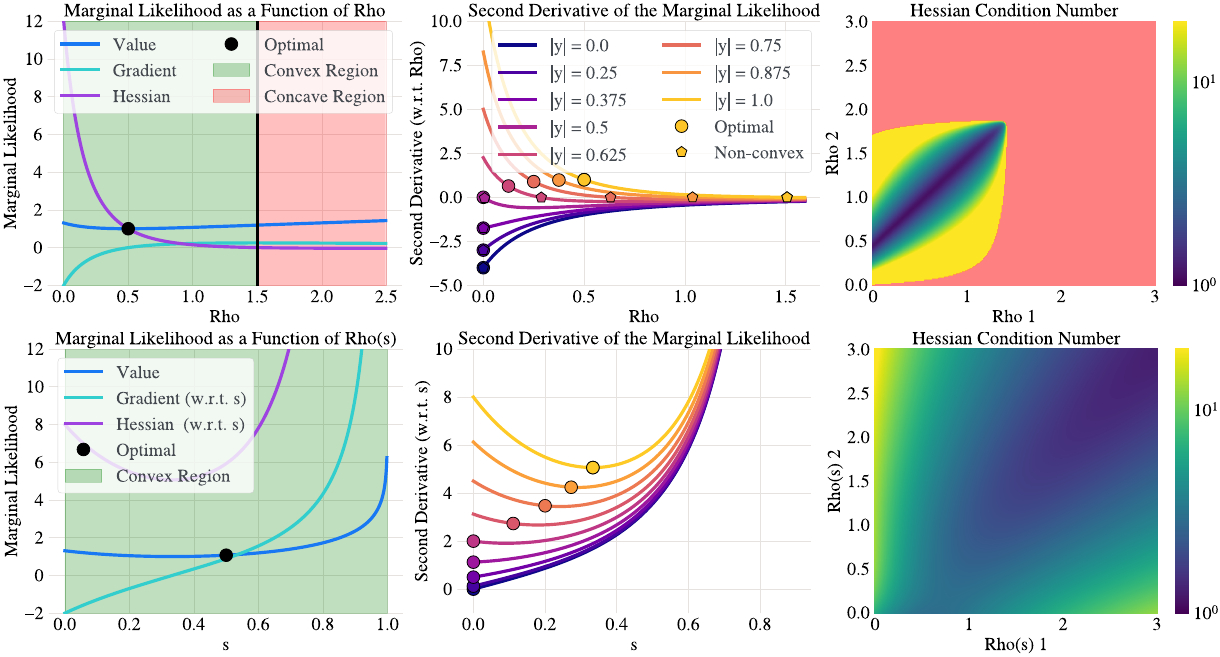}
    \caption{\textit{Top:} The behavior of the $-\log\calL(\rho)$ with respect to the canonical parameterization of $\brho$. 
    \textit{Bottom:} The behavior of $-\log\calL(\rho(\bs))$, highlighting the convexity 
    property.
    \textit{Left:} The value, and first two derivatives of $-\log\calL$ for a 1d example.
    \textit{Center:} The second derivatives of a 1d $-\log\calL$ as a function of $|y|$.
    The $\bs$-parameterization is everwhere convex for all considered $|y|$, while the canonical $\brho$-parameterization is only convex around the origin and only for $|y| > 0.5$.
    \textit{Right:} The heatmaps highlight that the original parameterization is non-convex (red) for larger values of $\rho$, and quickly becomes ill-conditioned, whereas
    the parameterization $\brho(\bs)$
    is convex and much better conditioned.
    }
    \label{fig:theory_figure}
\end{figure}

The NMLL $\calL$ of a GP~\eqref{eq:Preliminaries:GPs:mll} is the sum of a convex function $(\cdot)^{-1}$
and a concave function $\log \det (\cdot)$ of $\K$,
and is therefore not generally convex as a function of the hyper-parameters~$\btheta$, including the robust variances $\brho$.
Here, we propose a re-parameterization that allows us to prove strong convexity guarantees of the associated NMLL. 
In particular, we let $\brho(\mathbf s) = \diag(\K_0) \odot ((1 - \mathbf s)^{-1} - 1)$, where $\K_0 := k(\X, \X) + \sigma^2 \I$ and the inverse is element-wise. 
Note that $\brho(\mathbf s)$ is a diffeomorphism that maps~$\mathbf s$ from the compact domain $\mathbf s \in [0, 1]^n$ to the entire range of $\brho \in [0, \infty]^n$. 

Henceforth, we refer to the original $\brho$ 
as the {\it canonical} or $\brho$-parameterization and the newly proposed $\brho(\bs)$ as the {\it convex} or $\bs$-parameterization.
Lemma~\ref{lem:Theory:Convexity:ReparamHessian} shows the Hessian of the $\bs$-parameterization.

\begin{restatable}{lem}{reparameterizedhessian}[Reparameterized Hessian]
\label{lem:Theory:Convexity:ReparamHessian} 
 Let 
 $\K_{\bs} = k(\X, \X) + \sigma^2 \I + \D_{\brho(\bs)}$, 
 $\hat \K_\bs = \diag(\K_\bs)^{-1/2} \K_\bs \diag(\K_\bs)^{-1/2}$,
 and $\hat \balpha = \hat \K_\bs^{-1} \diag(\K_\bs)^{-1/2} \y$. 
 Then
\[
\Hess_{\mathbf s}[-2 \calL(\brho(\bs)] = 
\D_{1-\bs}^{-1}
 \left[
 2 \left( 
    \hat \balpha \hat \balpha^\top \odot (\hat \K^{-1} - \I)
    \right)
    +
    2\diag(\hat \K^{-1}) - (\hat \K^{-1} \odot \hat \K^{-1})
\right] 
\D_{1-\bs}^{-1}.
\]
\end{restatable}


Based on this representation, we now derive conditions 
on the eigenvalues of $\hat \K$ that imply the $m$-strong convexity and $M$-smoothness of the NMLL.
\begin{restatable}{lem}{eigenconvexity}[Strong Convexity via Eigenvalue Condition]
\label{lem:Theory:Convexity:EvalCondition}
Let $\hat \K_\bs$ as in Lemma~\ref{lem:Theory:Convexity:ReparamHessian}. 
Then $\Hess_{\bs} \succ m$ if 
\begin{align}
\label{eq:Theory:Convexity:EvalCondition:condition}
    \lambda_{\min} \hat\lambda_{\min}^2  
    \frac{(2 \hat \lambda_{\max}^{-1} - \hat \lambda_{\min}^{-2} - m)}{2(1 -  \lambda_{\min} / \lambda_{\max})} > \| \y \|_2^2,
\end{align}
where $\lambda_{\min, \max}$ (resp. $\hat \lambda_{\min, \max}$) are the smallest and largest eigenvalues of $ \K_\bs$, respectively $\hat \K_\bs$. 
\end{restatable}

The behavior
Lemma~\ref{lem:Theory:Convexity:EvalCondition} predicts is surprising 
and validated in Fig.~\ref{fig:theory_figure}.
Notably, the denominator ``blows up'' as $\K$ becomes close to unitary,
making the inequality more likely to be satisfied,
an indication that the convexity property of the NMLL 
is intimately linked to the RIP (Def.~\ref{defi:rip}).
Note that Lemma~\ref{lem:Theory:Convexity:EvalCondition}
is a condition for non-support-restricted convexity,
which is stronger than is necessary for the approximation guarantees
that rely on restricted convexity (Def.~\ref{defi:rsc}).
However, sparse eigenvalues are generally difficult to compute exactly. Fortunately, covariance matrices of GPs naturally tend to exhibit a property that
facilitates a different sufficient condition for convexity for 
all $\bs \in [0, 1]^{n}$.

\begin{defi}[Diagonal Dominance]
    A matrix $\mathbf A$ is said to be $\delta$-diagonally dominant if the elements $a_{ij}$ satisfy $\sum_{i \neq j} |a_{ij}| < \delta |a_{ii}|$ for all $i$.
\end{defi}


Intuitively, the $\rho_i(s)$ that are selected to be non-zero by the greedy algorithm take on large values,
further encouraging the diagonal dominance of 
the sub-matrix of $\K$ associated with the support of $\brho$.
For this reason, the following condition on $\K_0$ is 
sufficient to guarantee convexity for all $\bs \in [0, 1]^n$.

\begin{restatable}{lem}{ddconcavity}[Strong Convexity via Diagonal Dominance]
\label{lem:strong_convexity_via_diagonal_dominance}
    Let $m > 0$ and $\K_0$ be $\delta$-diagonally dominant with $\delta < 
    \left( (5 - m) -  \sqrt{25 - 9m + 17}\right) / 4
    \leq 
    (5 - \sqrt{17}) / 4 \approx 0.44$ and 
    \[
    \lambda_{\min}(\K_0) (1 - \delta)^2 
    \frac{2 (1 + \delta)^{-1} - (1-\delta)^{-2} - m}{2(1 -  (1 - \delta) / (1 + \delta))}
    \geq 
    \| \y \|^2_2.
    \]
    Then the NMLL is $m$-strongly convex for all $\bs \in [0,1]^n$, i.e. $\brho(\mathbf s) \in [0, \infty]^n$.
\end{restatable}
 
We attain similar results for $M$-smoothness,
see 
Lemma~\ref{lem:Theory:Smoothness:EvalCondition} and Lemma~\ref{lem:smoothness_via_diagonal_dominance} in the Appendix.
Having proven $m$-convexity and $M$-smoothness conditions,
we appeal to the results 
of \citet{elenberg2018restricted}.

\begin{restatable}{thm}{approxguarantee}[Approximation Guarantee]
    Let $\K_0 = k(\X, \X) + \sigma^2 \I$ be $\delta$-diagonally dominant,
    $s_{\max} > 0$ be an upper bound on $\|\bs\|_\infty$,
    and suppose $\|\y\|, \delta$ satisfy the bounds 
    of Lemmas~\ref{lem:strong_convexity_via_diagonal_dominance} and \ref{lem:smoothness_via_diagonal_dominance}, guaranteeing $m$-convexity and $M$-smoothness of the NMLL for some $m > 0$, $M > 1 / (1 - s_{\max})^2$.
    Let $\bs_{\operatorname{OMP}}(r)$ 
    be the $r$-sparse vector
    attained by OMP
    on the NMLL objective for $r$ steps,
    and let $\bs_{\operatorname{OPT}}(r) = \arg \max_{\|\bs\|_0=r, \|\bs\|_\infty \leq s_{\max}} \calL(\brho(\bs))$ be the optimal $r$-sparse vector.
    Then for any $2r \leq n$,
    \[
    \tilde \calL\left(\brho(\bs_{\operatorname{OMP}}(r))\right) 
    \ \geq \ 
    \bigl(1 - e^{-m / M} \bigr)
    \ 
    \tilde\calL\left(\brho(\bs_{\operatorname{OPT}}(r))\right),
    \]
where $\tilde \calL(\cdot) = \calL(\cdot) - \calL(\mathbf 0)$
is normalized so that
$\max_{\bs_\support} \tilde \calL(\bs_{\support}) \geq 0$ 
for any support $\support$.
\end{restatable}

A limitation of the theory is that it assumes the other hyper-parameters of the GP model to be constant,
as doing otherwise would introduce the non-convexity that is common to most marginal likelihood optimization problems.
In practice, we typically optimize $\brho$ jointly with the other hyper-parameters of the model in each iteration of \ouralgo{}, as this yields improved performance, see App.~\ref{subsec:convex_parameterization_for_joint_optimization} for details.

\section{Empirical Results}
\label{sec:EmpiricalResults}
We evaluate the empirical performance of \ouralgo{} 
against various baselines on a number of regression and Bayesian Optimization problems.
Specifically, we compare against a standard GP with a Matern-5/2 kernel (``Standard GP''), data pre-processing through Ax's adaptive winsorization procedure (``Adapt. Wins.'')~\citep{bakshy2018ae}, and a power transformation (``Power Transf.'')~\cite{box1964analysis}. 
Further, we also consider a Student-$t$ likelihood model from~\citet{jylanki2011robust} (``Student-$t$''), the trimmed marginal likelihood model from~\citet{andrade2023trimmed} (``Trimmed MLL''), and the RCGP model from \citet{altamirano2023robust}. Unless stated otherwise, all models are implemented in GPyTorch~\cite{gardner2018gpytorch} and all experiments in this section use $32$ replications.  See Appendix~\ref{appdx:AddEmpirical} for additional details.

\subsection{Regression Problems}
\label{subsec:EmpiricalResults:Regression}

\label{subsubsec:EmpiricalResults:Regression:Synthetic}
\paragraph{Synthetic}

\begin{figure}[!ht]
    \centering
    \includegraphics[width=\textwidth]{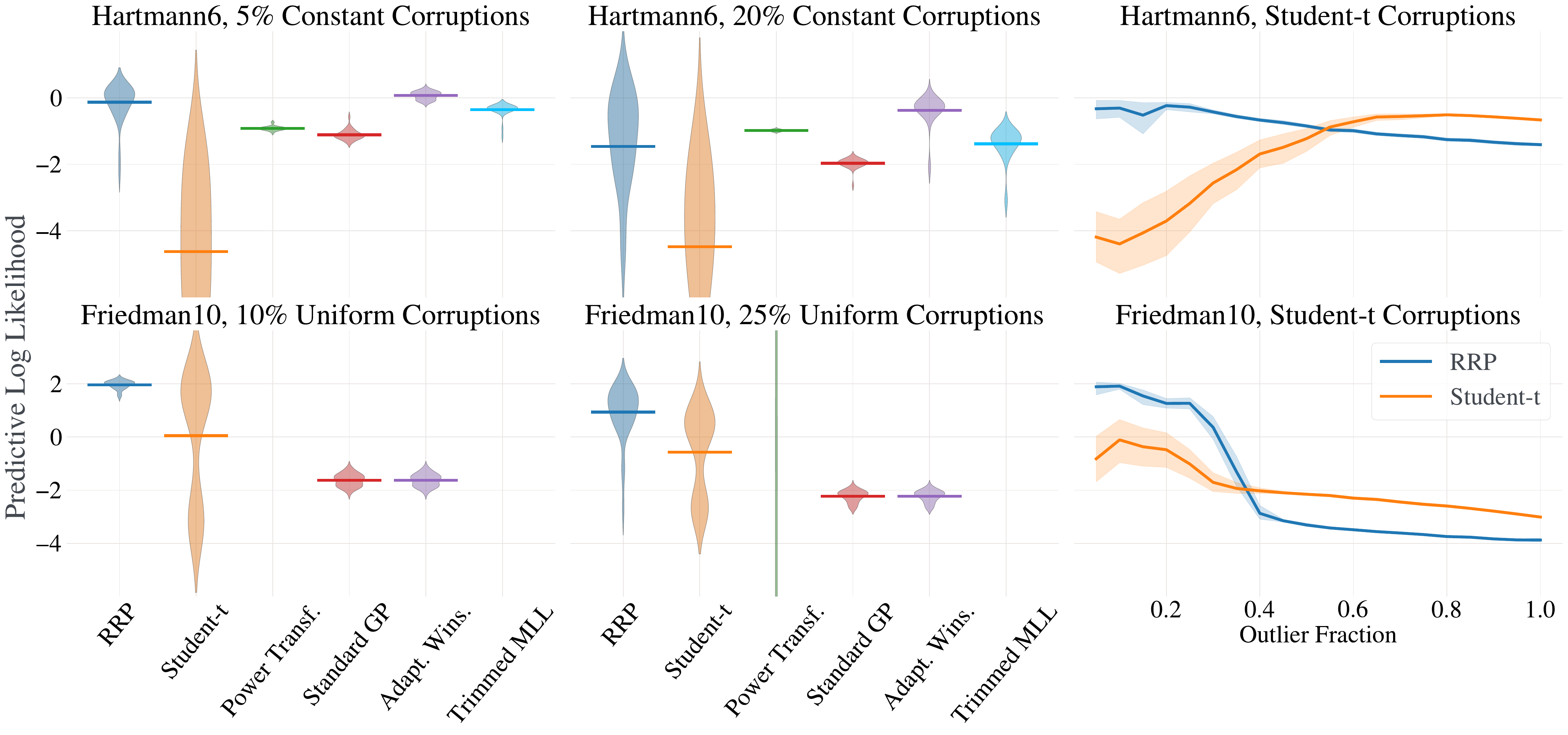}\\[-1.5ex]
    \caption{
    \textit{Left:} Distribution of predictive test-set log likelihood for various methods. Methods ommitted are those that performed substantially worse.
    \textit{Right:} Predictive log likelihood as a function 
    of the corruption probability for Student-$t$-distributed corruptions with two degrees of freedom.
    The GP model with the Student-$t$ likelihood 
    only starts outperforming \ouralgo{} as the corruption probability
    increases beyond 40\%, and exhibits a large variance in outcomes, which shrinks as the proportion of corruptions increases. All methods not shown were inferior to either \ouralgo{} or Student-$t$.
    }
\label{fig:EmpiricalResults:Regression:synthetic_violin}
\end{figure}

We first consider the popular Friedman10 and Hartmann6 \citep{dixon1978global} test functions from the literature.
We use two data generating processes: uniform noise, extreme outliers at some fixed value, and heavy-tailed (Student-$t$) noise at true function values. 
In these experiments, we compare the performance predictive log-likelihood. 
The results are shown in Fig.~\ref{fig:EmpiricalResults:Regression:synthetic_violin}.

\paragraph{Twitter Flash Crash}

\begin{figure}[!ht]
    \centering
    \includegraphics[width=0.99\textwidth]{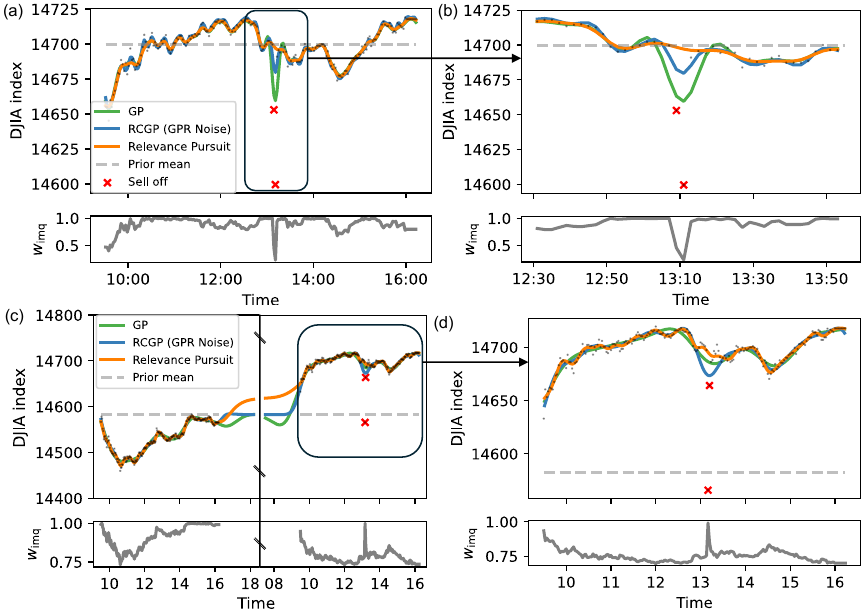}
    \caption{
        {
        Results on the intra-day data from the Dow Jones Industrial Average (DJIA) index on April 22-23 2013, which includes a sharp drop at 13:10 on the 23rd, see (b) for a detailed view.
        The accompanying panels labeled $w_{\text{imq}}$ show the
        function that \citet{altamirano2023robust}'s RCGP uses to down-weight data points. 
        \textit{Top}: RCGP, exhibits higher robustness than the standard GP, but is still affected by the outliers.
        The RRP model is virtually unaffected.
        \textit{Bottom}:
        Including the previous trading day into the training data in (c),
        leads RCGP to assign the {\it highest} weight $w_{\text{imq}}$ to the outlying data points due to their proximity to the target values' median,
        thereby leading RCGP to be even more affected than a standard GP,
        see (d) for a detailed view of the results on the data of April 23.
        }
    }
    \label{fig:EmpiricalResults:Regression:twitter}
\end{figure}

In Fig.~\ref{fig:EmpiricalResults:Regression:twitter},
we report a comparison to \citet{altamirano2023robust}'s RCGP
on data from the Dow Jones Industrial Average (DJIA) index on April 22-23 2013, which includes a sharp drop at 13:10 on the 23rd.
The top panels shows that RCGP exhibits higher robustness than the standard GP, but is still affected by the outliers, when trained on data from the 23rd. RRP is virtually unaffected.
Notably, RCGP relies on an a-priori weighting of data points based on the target values' proximity to their median, 
which can be counter-productive when the outliers are not a-priori separated in the range.
To highlight this, we included the previous trading day into the training data
for the bottom panels,
leading RCGP to assign the {\it highest} weight to the outlying data points due to their proximity to the target values' median,
thereby leading RCGP to ``trust'' the outliers more than any inlier,
resulting in it being less robust than a standard GP in this scenario.
See Appendix~\ref{appdx:AddEmpirical:rcgp} for additional comparisons to RCGP, 
on data sets from the UCI machine learning repository~\citep{kelly2023uci}.


\subsection{Robust Bayesian Optimization}
\label{subsec:EmpiricalResults:BayesOpt}
GPs are commonly used for Bayesian optimization (BO), which is a popular approach to sample-efficient black-box optimization~\cite{frazier2018tutorial}.
However, many of the GP models used for BO are sensitive to outliers and may not perform well in settings where such outliers occur. 
While~\citet{martinezcantin2017robust} consider the use of a Student-$t$ likelihood for BO with outliers, the use of other robust GP models has not been thoroughly studied in the literature.

\paragraph{Experimental setup}
We use \citet{ament2023logei}'s 
\href{https://github.com/pytorch/botorch/blob/66660e341b7dd0780feac4640f3709a8fd024206/botorch/acquisition/logei.py#L239C7-L239C35}{\texttt{qLogNoisyExpectedImprovement}}
(\texttt{qLogNEI}), a variant of the LogEI family of acquisition functions,
32 replications, and initialize all methods with the same quasi-random Sobol batch for each replication. 
We follow~\citet{hvarfner2024self} and plot the true value of the best in-sample point according to the GP model posterior at each iteration.
We also include Sobol and an ``Oracle'', which is a Standard GP that always observes the uncorrupted value,
and consider the backward canonical version of relevance pursuit, denoted by RRP, for these experiments.
The plots show the mean performance with a bootstrapped 90\% confidence interval.

\paragraph{Synthetic problems}
We consider the popular 6-dimensional Hartmann test function with three different corruption settings: (1) a constant value of $100$, (2) a $U[-3, 3]$ distributed value, (3) the objective value for a randomly chosen point in the domain.
The results for a 10\% corruption probability are shown in Fig.~\ref{fig:bo_hartmann}.
We also include results for a 20\% corruption probability in Appendix~\ref{appdx:additional_bo}.

\begin{figure}[!ht]
\centering
\includegraphics[width=\textwidth]{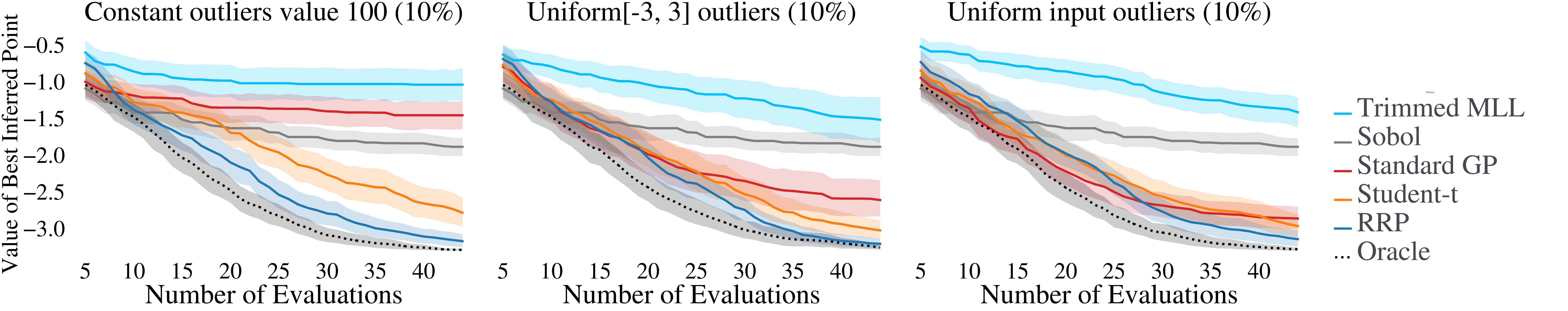}\\[-0.5ex] 
\caption{
    BO results for Hartmann6: 
    \textit{Left:} Relevance pursuit performs well in the case of constant outliers of value $100$, almost as well as the oracle. 
    \textit{Middle:} Relevance pursuit performs the best followed by the Student-$t$ likelihood in the case of $U[-3, 3]$. 
    \textit{Right:} Similar to the middle plot, this setting hides the corruptions within the range of the function, making it a challenging task.}
\label{fig:bo_hartmann}
\end{figure}

\paragraph{Real-world problems}
We include three real-world problems: 
A 3D SVM problem, a 5D CNN problem, and a 20D rover trajectory planning problem, see the App.~\ref{appdx:AddEmpirical:bo_details} for details.
For SVM and CNN, we simulate random corruptions corresponding to an I/O error, which causes the corresponding ML model to be trained using only a small subset of the training data.
For the rover 
planning 
problem we follow the setup in~\citep{robot} with the main difference that we consider a $20$D trajectory,
and the corruptions are generated randomly, causing the rover to break down at an arbitrary point along its trajectory.
In most cases, this results in a smaller reward than the reward of the full trajectory.

\begin{figure}[!ht]
\centering
\includegraphics[width=\textwidth]{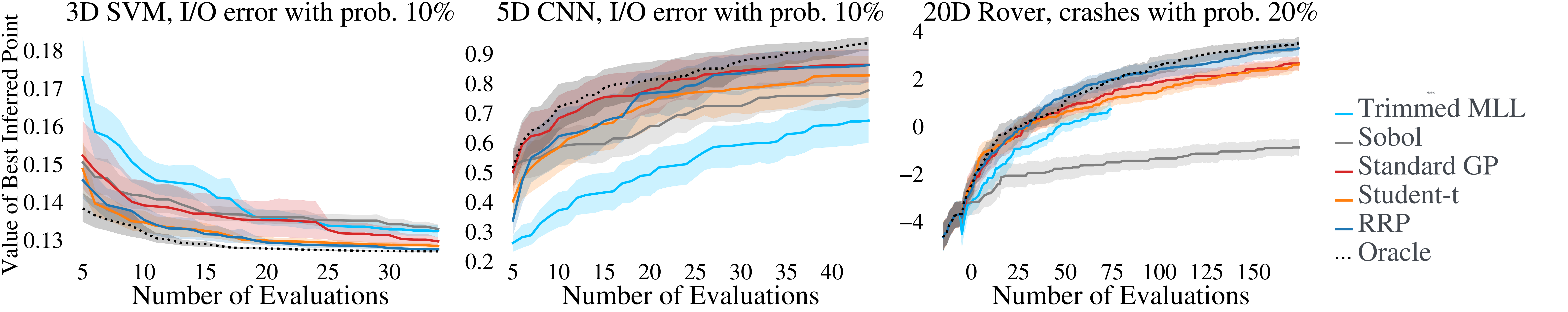}\\[-0.5ex] 
\caption{BO results for three real-world problems: 
\textit{Left:} \ouralgo{} is competitive with the oracle on the 3D SVM problem. 
\textit{Middle:} The power transform performs best on the 5D CNN problem, outperforming \ouralgo{} as well as the Oracle.
\textit{Right:} \ouralgo{} performs well on the 20D Rover problem.
}
\label{fig:bo_real_world}
\end{figure}

\section{Conclusion and Future Work}
\label{sec:Conclusion}

\paragraph{Contributions} Robust Gaussian Processes via Relevance Pursuit (\ouralgo{}) provides a novel and principled way to perform robust GP regression.
It permits efficient and robust inference, performs well across a variety of label corruption settings, retains good performance in the absence of corruptions, and is flexible, e.g., can be used with any mean or kernel function. 
Our method can be readily applied to both robust regression problems as well as applications such as Bayesian optimization
and is available through \href{https://botorch.org/}{\texttt{BoTorch}}~\citep{balandat2020botorch}.
Importantly, it also provides theoretical approximation guarantees.  

\paragraph{Limitations} As our approach does not explicitly consider the locations of the data points in the outlier identification, it may be outperformed by other methods if the underlying noise is heteroskedastic and location-dependent. 
On the other hand, those methods generally do not perform well in the presence of sparse, location-independent data corruptions. 

\paragraph{Extensions}
Promising extensions of this work include
performing Bayesian model averaging, i.e., average the predictions of the different possible sparsity models according to their likelihoods instead of using a MAP estimate,
applying \ouralgo{} to 
specialized models such as
\citet{lin2024scalinggaussianprocesseslearning}'s 
scalable learning-curve model for AutoML applications, 
and
\citet{ament2023sustainableconcretebayesianoptimization}'s
model for sustainable concrete.
On a higher level, the approach of combining greedy optimization algorithms with Bayesian model selection and leveraging a convex parameterization to achieve approximation guarantees 
might apply to other parameters that are optimized using the MLL objective: 
length-scales of stationary kernels,
coefficients of additive kernels, 
inducing inputs,
and even related model classes like 
\citet{tipping2001sparse}'s Sparse Bayesian Learning (SBL),
which seeks to identify sparse linear models
and is intimately linked to greedy matching pursuits~\citep{ament2021sblviastepwise}.
Overall, the approach has the potential to lead to theoretical guarantees, new insights, and performance improvements to widely-adopted Bayesian models.



\newpage
\bibliographystyle{abbrvnat}
\bibliography{main}

\clearpage
\appendix

\section{Additional Details on the Model}
\label{appdx:AddModel}

Algorithm~\ref{algo:RobustRegressionModel:Inference:RelevancePursuitBackward} below is the ``backward'' variant of Algorithm~\ref{algo:RobustRegressionModel:Inference:RelevancePursuitForward} from Sec~\ref{sec:rgp-rp}. As its name suggests, the main difference compared to the ``forward'' variant is that rather than building up a set of ``outliers'', it starts from a (typically large) set of ``outliers'' and iteratively \emph{removes} those data points from the set that have the smallest inferred data-point-dependent noise variance $\rho_i$. 

While we have not derived theoretical guarantees for this ``backward'' version, we have found it to generally behave similarly to the ``forward'' version in terms of performance and robustness. One empirical observation from our studies is that while the ``forward'' version tends to perform slightly better than the ``backward'' version if there are only few outliers, the opposite is true if the outlier frequency is very high. This behavior is rather intuitive and illustrates that relevance pursuit is particularly well-suited to identify sparse, low-cardinality subsets (note that in the ``backward'' variant under large corruptions, the uncorrupted data points can be viewed as the sparse subset that needs to be identified).

\begin{algorithm}
\caption{Relevance Pursuit (Backward Algorithm)}
\label{algo:RobustRegressionModel:Inference:RelevancePursuitBackward}
\begin{algorithmic}
    \Require $\X$, $\y$, schedule $\calK =(k_1, k_2, \dotsc )$
    \State Initialize $\support_0^c \subseteq \{1, \hdots, n\}$ (typically $\support_0^c = \{1, \hdots, n\})$
    \For{$k_i$ in $\calK$}
        \State Optimize ML: $\brho_{\support_i^c}^* \leftarrow \arg \max_{\brho_{\support_i^c}} 
        \calL(\brho_{\support_i^c})$, \; where $\brho_{\support_i^c} = \{\brho: \rho_j = 0, \; \forall \, j \notin \support_i^c \} $
        \State Compute the set $\calR_i$ containing the $k_i$ elements of $\support_i^c$ with smallest inferred variance:
        \State \qquad $\calR_i \leftarrow \{j_i^1, \dotsc, j_i^{k_i}\}$ where $j_i^l \in \support_i^c$ such that $\rho_{\support_i^c}^*(j_i^l) \leq \rho_{\support_i^c}^*(j_i^{l'})$ for $l<l'$
        \State $\support_{i+1}^c \leftarrow \support_i^c \setminus \calR_i$
    \EndFor
    \State $\support_i \leftarrow \{1, \hdots, n\} \setminus \support_i^c$ for each $k_i$
    \State Compute the marginal likelihood $p(\support_i | \X, \y) \approx p(\support_i, \brho_{\support_i}^*, | \X, \y)$
    \State $\support^* \leftarrow \arg \max_{\support_i} p(\support_i | \X, \y) p(\support_i)$.
    \State Return $\support^*$, $\brho_{\support^*}^*$.
\end{algorithmic}
\end{algorithm}

\section{Additional Background on the Theory}
\label{appdx:AddTheory}

\subsection{Submodular Functions}
\label{appdx:AddTheory:Submodular}

\citet{krause2014submodularsurvey} provides a survey on the maximization of general submodular functions. Here, we focus on applications of submodularity to sparse regression and Gaussian process models.

\paragraph{Sparse Regression} \citet{das2018approximate} showed that the subset selection problem of regression features with an $R^2$ objective 
satisfies a weak submodularity property,
which can be invoked to prove approximation guarantees for the greedy maximization of the objective.
\citet{elenberg2018restricted} generalized this work
by proving that any log likelihood function exhibiting restricted strong concavity gives rise to the weak submodularity of the associated subset selection problem, 
which can be invoked to prove approximation guarantees for the greedy algorithm.
\citet{karaca2021performance} contains a guarantee for the backward algorithm applied to the maximization of submodular set functions.


\paragraph{Gaussian Processes}
Submodularity has also found application to Gaussian process models.
For a sensor placement problem,
\citet{krause2008near} proved that the mutual information (MI) criterion,
capturing the reduction in uncertainty in the entire search space,
can be a submodular function.
In this case, MI is not monotonic everywhere,
but monotonic for small sets ($2k$) of sensors,
which is sufficient to apply Nemhauser's guarantee 
for sparse sets of sensors up to size $k$~\citep{nemhauser1978analysis}.
Relatedly, the ``myopic'' joint entropy of a set of observables is unconditionally submodular as a consequence of the ``information never hurts'' principle~\citep{krause2005near},
but generally leads to lower-quality sensor placements than the MI criterion.
\citet{srinivas2010regret} used the submodularity of the joint entropy in order to prove regret bounds for the convergence of a GP-based BO algorithm using the upper-confidence bound acquisition function.

\citet{elenberg2018restricted} proved that any log likelihood function exhibiting restricted strong concavity and smoothness implies the weak submodularity of the associated subset selection problem. 
\begin{defi}[Submodularity Ratios \citep{elenberg2018restricted}]
    Let $\mathcal{A}, \mathcal{B} \subset [n]$ be two disjoint sets,
    and $f : 2^{[n]} \to \R$.
    The submodularity ratio of $\mathcal{B}$ with respect to $\mathcal{A}$
    is defined by 
    \[
    \gamma_{\mathcal{B}, \mathcal{A}} = 
        \sum_{i \in \mathcal{A}} \left(
            f(\mathcal{B} \cup \{i\}) - f(\mathcal{B}) 
        \right)
        \big /
        (f(\mathcal{B} \cup \mathcal{A}) - f(\mathcal{B}))
    \]
    The submodularity ratio of a set $\mathcal{C}$ with respect to an integer $k$ is defined by 
    \[
    \gamma_{\mathcal{C}, k} = \min_{\mathcal{B}, \mathcal{A}} \gamma_{\mathcal{B}, \mathcal{A}} 
    \qquad \text{such that} \qquad 
    \mathcal{A} \cap \mathcal{B} = \emptyset, \qquad
    \mathcal{B} \subseteq \mathcal{C}, \qquad \text{and} \qquad
    |\mathcal{A}| \leq k.
    \]
    Then given $\gamma > 0$, 
    a function is $\gamma$-weakly submodular at a set $\mathcal{C}$
    with respect to $k$ if $\gamma_{\mathcal{C}, k} \geq \gamma$.
\end{defi}

\begin{thm}[Weak Submodularity via RSC \citep{elenberg2018restricted}]
The submodularity ratio $\gamma_{\support, k}$ can be bound below using the
restricted convexity and smoothness parameters
$m_{|\support| + k}$ and $\ M_{|\support | + k}$,
    \[
    \gamma_{\support, k} \geq m_{|\support| + k} \ \big / \ M_{|\support | + k}.
    \]
\end{thm}

\begin{thm}[OMP Approximation Guarantee \citep{elenberg2018restricted}]
\label{thm:omp_approximation_guarantee}
Let $\x_{\operatorname{OMP}}(r)$
be the $r$-sparse vector selected by OMP,
and $\x_{\operatorname{OPT}}(r) = \arg \min_{\|\x\|_0=r} f(\x)$
be the optimal $r$-sparse vector.
Then
\[
f\left(\x_{\operatorname{OMP}}(r)\right) \geq \left(1 - e^{-m_{2r} / M_{2r}} \right) f\left(\x_{\operatorname{OPT}}(r)\right),
\]
where $m_{2r}, M_{2r}$ are the restricted strong convexity and smoothness parameters of $f$, respectively.
\end{thm}

\section{Theoretical Results and Proofs}
\label{appdx:proofs}

\robustvariance*
\begin{proof}
First, we partition the covariance matrix $\K + \sigma^2 \I + \D_{\rho}$ 
to separate the effect of $\rho_i$ and use the block matrix inverse
    \begin{equation}
        \begin{aligned}
            \bSigma^{-1} = (\K + \D_{\brho + \sigma^2})^{-1}
                &= \begin{bmatrix}
                    \bSigma_{\backslash i}^{-1} + \bu \beta_i \bu^\top 
                    & -\bu \beta_i \\
                    - \bu^\top \beta_i 
                    & \beta_i \\ 
                \end{bmatrix},
        \end{aligned}
    \end{equation}
    where 
    \begin{equation}
        \begin{aligned}
            \bSigma_{\backslash i} &= k(\X_{\backslash i}, \X_{\backslash i}) + \D_{\brho_{\backslash i} + \sigma^2}, \\ 
            \bu &= \bSigma_{\backslash i}^{-1} k(\X_{\backslash i}, \x_i), 
            \qquad \text{and} \\
            \beta_i &= \left( 
                [k(\x_i, \x_i) + \sigma^2 + \rho_i] - k(\x_i, \X_{\backslash i}) \bSigma_{\backslash i}^{-1} k(\X_{\backslash i}, \x_i)
                \right)^{-1}.
        \end{aligned}
    \end{equation}
    \paragraph{Quadratic Term}
    With the expression for the inverse of $\bSigma$ above, we can write the
    quadratic term of the log likelihood as
    \begin{equation}
        \begin{aligned}
            \y^\top (\K + \D_{\brho + \sigma^2})^{-1} \y
                &= \y_{\backslash i}^\top (\bSigma_{\backslash i}^{-1} + \bu \beta_i \bu^\top) \y_{\backslash i}
                - 2 \y_{\backslash i}^\top \bu \beta_i y_i 
                + y_i^2 \beta_i \\
                &= \y_{\backslash i}^\top \bSigma_{\backslash i}^{-1} \y_{\backslash i} + \beta_i (\y_{\backslash i}^\top \bu - y_i)^2.
        \end{aligned}
    \end{equation}

    \paragraph{Determinant Term}
    The determinant of a block matrix is given by 
    \begin{equation}
        \left| \begin{bmatrix}
            \mathbf A & \mathbf B \\
            \mathbf C & \mathbf D \\
        \end{bmatrix} \right|
        = |\mathbf A| |\mathbf D - \mathbf C \mathbf A^{-1} \mathbf B|.
    \end{equation}
    Applying this identity to $\bSigma = (\K + \D_{\brho + \sigma^2})$, we get
    \begin{equation}
        \begin{aligned}
            |\K + \D_{\brho + \sigma^2}| = |\bSigma_{\backslash i}| \beta_i^{-1}.
        \end{aligned}
    \end{equation}

    \paragraph{Log Marginal Likelihood}
    \begin{equation}
        2(\calL(\brho) - \calL(\brho_{\backslash i})) = - \beta_i (\y_{\backslash i}^\top \bu - y_i)^2 - \log(\beta_i^{-1}).
    \end{equation}
    Noting that $\partial_{\rho_i} \beta_i = -\beta_i^2$,
    the derivative of the difference in log marginal likelihood w.r.t. $\rho_i$, is
    \begin{equation}
        \partial_{\rho_i} 2(\calL(\brho) - \calL(\brho_{\backslash i})) 
        = (\y_{\backslash i}^\top \bu - y_i)^2 \beta_i^2 - \beta_i.
    \end{equation}
    While $\beta_i = 0$ is a root of the derivative, we ignore this solution
    since $\beta_i$ is never zero when $\sigma^2 > 0$.
    Therefore, the remaining stationary point is $\beta_i^{-1} = (\y_{\backslash i}^\top \bu - y_i)^2$.
    Since we constrain $\rho \geq 0$, this point might not always be attainable.
    However, because there is only a single stationary point with respect to $\beta_i$ when $\sigma^2 > 0$,
    and $\beta_i$ is a strictly decreasing function of $\rho_i$,
    it follows that the marginal likelihood is monotonic as a function of $\rho_i$ to both the left and the right of the stationary point.
    Therefore, the optimal constraint $\rho_i$ is simply the optimal unconstrained value, projected into the feasible space. In particular,
    solving $\beta_i^{-1} = (\y_{\backslash i}^\top \bu - y_i)^2$ for $\rho_i$ and projecting to the non-negative half-line, we get
    \begin{equation}
        \rho_i = \left[
            \left(
                \y_{\backslash i}^\top \bu - y_i
            \right)^2 
            - 
            \left(
                k(\x_i, \x_i) + \sigma^2 - k(\x_i, \X_{\backslash i}) \bSigma_{\backslash i}^{-1} k(\X_{\backslash i}, \x_i)
            \right) 
        \right]_{+}.
    \end{equation}
    Lastly, note
    \begin{equation}
    \begin{aligned}
        \y_{\backslash i}^\top \bu - y_i = \y_{\backslash i}^\top (k(\X_{\backslash i}, \X_{\backslash i}) + \D_{\sigma^2 + \brho_{\backslash i}})^{-1} k(\X_{\backslash i}, \x_i) - y_i 
        &= \E[y(\x_i) | \calD_{\backslash i}] - y_i, \\
        k(\x_i, \x_i) - k(\x_i, \X_{\backslash i}) \bSigma_{\backslash i}^{-1} k(\X_{\backslash i}, \x_i) + \sigma^2
        &= \V[y(\x_i) | \calD_{\backslash i}].
    \end{aligned}
    \end{equation}
    
    As stated by~\citet{rasmussen2006gaussian} (P. 117, Eq. 5.12),
    originally shown by \citet{sundararajan1999optloo},
    these quantities can be expressed as simple functions of $\bSigma^{-1}$:
    \begin{equation}
        \begin{aligned}
            \E[y(\x_i) | \calD_{\backslash i}]^2 &= y_i - \left[
                \bSigma^{-1} \y
                \right]_i
                \big / \left[ \bSigma^{-1} \right]_{ii} \\ 
            \V[y(\x_i) | \calD_{\backslash i}] &= 1 \big / \left[ \bSigma^{-1} \right]_{ii}.
        \end{aligned}
    \end{equation} 
    Therefore, all LOO predictive values can be computed in $\mathcal{O}(n^3)$
    or faster, if an inducing point method is used for $\K$.
\end{proof}

The following is a preliminary result for our analysis of the log marginal likelihood w.r.t. $\brho$.

\begin{restatable}{lem}{robustgradienthessian}
\label{lem:robust_gradient_hessian}
    The gradient and Hessian of the log marginal likelihood $\calL$ with respect to $\brho$ are given by
    \[
        -2 \nabla_{\brho}[\calL] = \diag(\K^{-1} - \balpha \balpha^\top), 
        \qquad \text{and} \qquad
        -2 \Hess_{\brho}[\calL] = (2\balpha \balpha^\top - \K^{-1}) \odot \K^{-1},
    \]
    where $\K = \K_0 + \D_{\brho}$ for some base covariance matrix $\K_0$ and $\balpha = \K^{-1} \y$.
\end{restatable}   
\begin{proof}
    Let $\balpha = \K^{-1} \y$. 
    Regarding the gradient,
    note that $\partial_{\rho_i} \K = \mathbf{e}_i\mathbf{e}_i^\top$,
where $\mathbf{e}_i$ is the canonical basis vector with a one as the $i$th element, and
    based on Equation~5.9 of \citet{rasmussen2006gaussian},
    \[
    \begin{aligned}
    -2 \partial_{\rho_i} [\calL] &= 
    \tr \left(
        \left(
            \K^{-1} - \balpha \balpha^\top
        \right) \partial_{\rho_i} \K
    \right) \\
    &= 
    \tr(
        \left(
            \K^{-1} - \balpha \balpha^\top
        \right) \mathbf e_i \mathbf e_i^\top
    ) \\
    &= 
        \mathbf e_i^\top\left(
            \K^{-1} - \balpha \balpha^\top
        \right) \mathbf e_i \\
    &= 
        \left(
            \K^{-1} - \balpha \balpha^\top
        \right)_{ii}. \\
    \end{aligned}
    \]
Regarding the second derivatives, according to \citet{dong2017scalable},
\[
\begin{aligned}
    \partial_{\theta_i} \partial_{\theta_j} [\log |\K|]
    &= \tr(
        \K^{-1} [\partial_{\theta_i} \partial_{\theta_j} \K] - \K^{-1} [\partial_{\theta_i} \K] \K^{-1} [\partial_{\theta_j} \K]
    ) \\ 
\partial_{\theta_i} \partial_{\theta_j} [\y^\top \K^{-1} \y]
    &= 2\balpha^T [\partial_{\theta_i} \K]
    \K^{-1}
    [\partial_{\theta_j} \K] \balpha
    - \balpha^\top 
    [\partial_{\theta_i} \partial_{\theta_j} \K]
    \balpha.
\end{aligned}
\]
Therefore,
\[
\begin{aligned}
   -2 \partial_{\rho_i} \partial_{\rho_j} \calL &= 
    \tr \left(
        (\K^{-1} - \balpha \balpha^T) [\partial_{\rho_i} \partial_{\rho_j} \K]
        - (\K^{-1} - 2 \balpha \balpha^\top) 
        ([\partial_{\rho_i} \K] \K^{-1} [\partial_{\rho_j} \K])
    \right) \\
    &= \tr \left(
        (2 \balpha \balpha^\top - \K^{-1}) 
        ([\K^{-1}]_{ij} \mathbf{e}_i \mathbf{e}_j^\top)
    \right) \\
    &= 
    \tr \left(
        \mathbf{e}_j^\top (2 \balpha \balpha^\top - \K^{-1}) 
        \mathbf{e}_i
    \right) [\K^{-1}]_{ij} \\
    &= [2 \balpha \balpha^\top - \K^{-1}]_{ij} [\K^{-1}]_{ij} \\
    &= \left[ (2\balpha \balpha^\top - \K^{-1}) \odot \K^{-1} \right]_{ij},
\end{aligned}   
\]
since $[\partial_{\rho_i} \partial_{\rho_j} \K] = \mathbf 0$.
The third equality is due to the invariance of the trace to circular shifts of its argument.
The forth equality is due to the symmetry of the matrix in brackets.
\end{proof}

\subsection{Strong Convexity and Smoothness of the Reparameterized Robust Marginal Likelihood}

Here, we re-parameterize $\mathbf \brho(\mathbf s) = \diag(\K_0) \odot ((1 - \bs)^{-1} - 1)$,
and attain strong convexity for all inputs $\mathbf s$,
if conditions on the eigenvalues of the covariance matrix
and the norm of the data vector $\|\y\|$ are met.
The convexity result is surprising in two ways:
the negative log marginal likelihood of GPs is generally a non-convex function,
and in addition,
the negative log likelihoods of many alternative robust regression methods 
like the Student-$t$ likelihood or $\alpha$-stable likelihoods are non-convex, and even Huber's proposal is non-{\it strongly}-convex.

\reparameterizedhessian*
\begin{proof}
Using the chain-rule, the Hessian $\Hess_\bs[\calL]$ can be expressed as 
a function of the Jacobian $\mathbf J_{\bs}[\brho(\bs)] = \D_{\partial \brho(\bs)}$, which is diagonal since $\brho(\bs)$ is an element-wise function,
and the second derivatives $\partial_s^2 \rho(s_i)$.
Then
\[
\begin{aligned}
    -2\Hess_\bs [\calL] 
    &= -{\mathbf J}_\bs [\brho]^\top \Hess_\brho[2\log \calL] {\mathbf J}_\bs [\brho]
    + \D_{\nabla_\brho [-2 \calL]} \D_{\partial_s^2 \brho} \\
    &=  \D_{\brho'(\bs)} [(2\balpha \balpha^\top - \K^{-1}) \odot \K^{-1}] \D_{\brho'(\bs)} 
    + \diag(\K^{-1}) \D_{\brho''} - \D_{\brho'} (\K^{-1} \circ \K^{-1}) \D_{\brho'},
\end{aligned}
\]
where we substituted the relevant expressions from Lemma~\ref{lem:robust_gradient_hessian}.
Further substituting $\brho(\mathbf s)_i = [\K_0]_{ii} [(1 - s_i)^{-1} - 1]$,
$\brho'(\mathbf s)_i = [\K_0]_{ii} (1 - s_i)^{-2}$,
and $\brho''(\mathbf s)_i = 2[\K_0]_{ii} (1 - s_i)^{-3}$,
noting that $\K = \K_0 + \diag(\K_0) [(1-s)^{-1} - 1] = (\K_0 - \diag(\K_0)) + \diag(\K_0) (1-s)^{-1}$,
and algebraic manipulation finish the proof.
\end{proof}

\eigenconvexity*
\begin{proof}
We seek to lower-bound
the smallest eigenvalue of the Hessian matrix, which---for twice-continuously-differentiable problems---is equivalent to lower and upper bounds of the problem's Hessian matrix.
Starting with the result of Lemma~\ref{lem:Theory:Convexity:ReparamHessian},
\[
\begin{aligned}
    \Hess_{\mathbf s}[-2 \calL(\brho(\bs)] 
    &= 
\D_{1-\bs}^{-1}
 \left[
 2 \left( 
    \hat \balpha \hat \balpha^\top \odot (\hat \K^{-1} - \I)
    \right)
    +
    2\diag(\hat \K^{-1}) - (\hat \K^{-1} \odot \hat \K^{-1})
\right] 
\D_{1-\bs}^{-1} \\
    &\succeq
    2 \left( 
    \hat \balpha \hat \balpha^\top \odot (\hat \K^{-1} - \I)
    \right)
    +
    2\diag(\hat \K^{-1}) - (\hat \K^{-1} \odot \hat \K^{-1}).
\end{aligned}
\]
Now, we bound each of the three additive terms independently 
from below.

{\bf Term 1:}
\[
\begin{aligned}
    2\diag(\hat \K^{-1}) 
    \succeq 2 \lambda_{\min}(\hat \K^{-1}) \I
    = 2 \lambda_{\max}(\hat \K)^{-1} \I.
\end{aligned}
\]
The first inequality comes from $\K$ being positive definite,
and the absolute value of the diagonal of a matrix,
which is already positive for positive definite matrices,
being lower bounded by the minimum eigenvalue of the matrix.
The last steps is a basic consequence of the eigenvalues
of inverses matrices. 
Note that the eigenvalues of $\hat \K$ can be further bound 
by the eigenvalues of the original matrix $\K$:
\[
\begin{aligned}
\lambda_{\min}(\hat \K) &= \lambda_{\min}(\diag(\K)^{-1/2} \K \diag(\K)^{-1/2})  \\
    &\geq \lambda_{\min}(\diag(\K)^{-1/2})^2 \lambda_{\min}(\K)  \\
    &= \lambda_{\min}(\diag(\K)^{-1}) \lambda_{\min}(\K) \\
    &\geq \lambda_{\min}(\K^{-1}) \lambda_{\min}(\K) \\
    &= \lambda_{\min}(\K) / \lambda_{\max}(\K).
\end{aligned}
\]
In a similar way, we can show that $\lambda_{\max}(\hat \K) \leq \lambda_{\max}(\K) / \lambda_{\min}(\K)$,
which implies 
$\lambda_{\max}(\hat \K)^{-1} \geq \lambda_{\min}(\K) / \lambda_{\max}(\K)$.

{\bf Term 2:} Next, we have
\[
\begin{aligned}
    - (\hat \K^{-1} \odot \hat \K^{-1}) 
    \succeq - \lambda_{\max}(\hat \K^{-1})^2 \I = 
    -  \lambda_{\min}(\hat \K)^{-2} \I,
\end{aligned}
\]
which is due to the Hadamard product being a sub-matrix of the Kronecker product 
of the same matrices, the largest eigenvalue of the former
are bounded by the largest eigenvalue of the latter,
which is the product of the largest eigenvalues of the constituent matrices.

{\bf Term 3:} Lastly, note that
\[
\begin{aligned}
2 \left( 
    \hat \balpha \hat \balpha^\top \odot (\hat \K^{-1} - \I)
    \right)
    &= 
    2\D_{\hat \balpha} (\hat \K^{-1} - \I) \D_{\hat \balpha} \\
    &\succeq 
    2\D_{\hat \balpha} (\lambda_{\min}(\hat \K^{-1})\I - \I) \D_{\hat \balpha} \\
    &= 
    2(\lambda_{\min}(\hat \K^{-1}) - 1) \D_{\hat \balpha}^2 \\
    &=
    2(\lambda_{\max}(\hat \K)^{-1} - 1) \D_{\hat \balpha}^2 \\
    &\succeq
    2(\lambda_{\min}(\K) / \lambda_{\max}(\K) - 1) \D_{\hat \balpha}^2 \\
    &\succeq
    2(\lambda_{\min}(\K) / \lambda_{\max}(\K) - 1) \|\hat \balpha\|_\infty^2 \I,
\end{aligned}
\]
where the last inequality comes from the second to last lower bound of Term 3 being non-positive, and therefore, being able to lower bound it with the largest magnitude entry of $\hat \balpha$.

{\bf Term 1 + 2 + 3:}
Putting together the inequalities for all terms, we get
\[
\begin{aligned}
\lambda_{\min}(\Hess_{\bs}) 
    &\geq 2 \lambda_{\max}(\hat \K)^{-1} - \lambda_{\min}(\hat \K)^{-2} 
    + 2(\lambda_{\min}(\K) / \lambda_{\max}(\K) - 1) \|\hat \balpha\|_\infty^2, \\  
\end{aligned}
\]
where in slight abuse of notation, we let $\Hess_\bs = \Hess_\bs[-2\calL]$ 
be the Hessian of the negative log likelihood.

Using the bound 
$\|\hat \balpha\|_\infty \leq \|\hat \balpha\|_2 \leq \lambda_{\min}(\hat \K)^{-1} \|\hat \y\|$,
where $\hat \y = \diag(\K)^{-1/2} \y$.
Therefore,
$\|\hat \balpha\|_\infty \leq \lambda_{\min}(\hat \K)^{-1} \lambda_{\min}(\K)^{-1/2} \|\y\|_2 \leq \lambda_{\max}(\K) \lambda_{\min}(\K)^{-3/2} \|\y\|_2$.

Finally, lower bounding the current lower bound by $m > 0$ yields a sufficient condition for the convexity at $\bs$.
\[
\begin{aligned}
    2 \lambda_{\max}(\hat \K)^{-1} - \lambda_{\min}(\hat \K)^{-2} 
    + 2(\lambda_{\min}(\K) / \lambda_{\max}(\K) - 1) \lambda_{\min}(\hat \K)^{-2} \lambda_{\min}(\K)^{-1} \|\y\|_2^2 > m. \\  
\end{aligned}
\]
Re-arranging, we attain
\begin{align}
    \lambda_{\min} \hat\lambda_{\min}^2  
    \frac{(2 \hat \lambda_{\max}^{-1} - \hat \lambda_{\min}^{-2} - m)}{2(1 -  \lambda_{\min} / \lambda_{\max})} > \| \y \|_2^2,
\end{align}
which is a non-trivial guarantee when 
$2 \hat \lambda_{\max}^{-1} - \hat \lambda_{\min}^{-2} - m > 0$.
\end{proof}

\begin{restatable}{lem}{eigensmoothness}[Smoothness via Eigenvalue Condition]
\label{lem:Theory:Smoothness:EvalCondition}
Let $\hat \K$ as in Lemma~\ref{lem:Theory:Convexity:ReparamHessian}. Suppose that $\|\bs\|_{\infty} \leq s_{\max}$ and
\begin{align}
\label{eq:Theory:Smoothness:EvalCondition:condition}
    \lambda_{\min} \hat\lambda_{\min}^2  
    \frac{(M (1 - s_{\max})^{2} + \hat \lambda_{\min}^{-2} - 2 \hat \lambda_{\max}^{-1})}{2(\lambda_{\max} / \lambda_{\min} - 1)} > \| \y \|_2^2,
\end{align}
where $\lambda_{\min, \max}$ (resp. $\hat \lambda_{\min, \max}$) are the smallest and largest eigenvalues of $ \K$ (resp. $\hat \K$) respectively. Then $\Hess_{\bs} \prec M$.
This is a non-trivial guarantee when 
$M (1 - s_{\max})^{2} + \hat \lambda_{\min}^{-2} - 2 \hat \lambda_{\max}^{-1} > 0$.
\end{restatable}
\begin{proof}
We now derive an equivalent upper bound for the largest eigenvalue of the Hessian. 
Starting with the result of Lemma~\ref{lem:Theory:Convexity:ReparamHessian},
\[
\begin{aligned}
    \Hess_{\mathbf s}[-2 \calL(\brho(\bs)] 
    &= 
\D_{1-\bs}^{-1}
 \left[
 2 \left( 
    \hat \balpha \hat \balpha^\top \odot (\hat \K^{-1} - \I)
    \right)
    +
    2\diag(\hat \K^{-1}) - (\hat \K^{-1} \odot \hat \K^{-1})
\right] 
\D_{1-\bs}^{-1} \\
    &\preceq
    \|(1 - \bs)^{-1}\|_\infty^2
     \left[
    2 \left( 
    \hat \balpha \hat \balpha^\top \odot (\hat \K^{-1} - \I)
    \right)
    +
    2\diag(\hat \K^{-1}) - (\hat \K^{-1} \odot \hat \K^{-1})
    \right].
\end{aligned}
\]
Therefore, it becomes immediately apparent that 
we will need to introduce an upper bound on $\|(1 - \bs)^{-1}\|_\infty^2$,
which is a restriction on the domain that $\brho$ can take.
Proceeding in a similar way as above, 
we bound the three terms in square brackets, 
now from above.

{\bf Term 1:}
\[
\begin{aligned}
    2\diag(\hat \K^{-1}) 
    \preceq 2 \lambda_{\max}(\hat \K^{-1}) \I
    = 2 \lambda_{\min}(\hat \K)^{-1} \I.
\end{aligned}
\]

{\bf Term 2:}
\[
\begin{aligned}
    - (\hat \K^{-1} \odot \hat \K^{-1}) 
    \preceq - \lambda_{\min}(\hat \K^{-1})^2 \I = 
    -  \lambda_{\max}(\hat \K)^{-2} \I.
\end{aligned}
\]
{\bf Term 3:}
\[
\begin{aligned}
2 \left( 
    \hat \balpha \hat \balpha^\top \odot (\hat \K^{-1} - \I)
    \right)
    &= 
    2\D_{\hat \balpha} (\hat \K^{-1} - \I) \D_{\hat \balpha} \\
    &\preceq 
    2\D_{\hat \balpha} (\lambda_{\max}(\hat \K^{-1})\I - \I) \D_{\hat \balpha} \\
    &= 
    2(\lambda_{\max}(\hat \K^{-1}) - 1) \D_{\hat \balpha}^2 \\
    &=
    2(\lambda_{\min}(\hat \K)^{-1} - 1) \D_{\hat \balpha}^2 \\
    &\preceq
    2(\lambda_{\max}(\K) / \lambda_{\min}(\K) - 1) \D_{\hat \balpha}^2 \\
    &\preceq
    2(\lambda_{\max}(\K) / \lambda_{\min}(\K) - 1) \|\hat \balpha\|_\infty^2 \I,
\end{aligned}
\]
where now the last inequality follows because the second to last expression is always non-negative.
{\bf Term 1 + 2 + 3:}
Putting together the inequalities for all terms, we get
\[
\begin{aligned}
\lambda_{\max}(\Hess_{\bs}) 
    &\leq 2 \lambda_{\min}(\hat \K)^{-1} - \lambda_{\max}(\hat \K)^{-2} 
    + 2(\lambda_{\max}(\K) / \lambda_{\min}(\K) - 1) \|\hat \balpha\|_\infty^2, \\  
\end{aligned}
\]
Finally, upper bounding the current upper bound by $M / \|(1 - \bs)^{-1}\|_\infty^2 > 0$ yields a sufficient condition for the $M$-smoothness at $\bs$.
Using the same bound for $\|\hat \balpha\|_\infty$ derived for the convexity result,
\[
\begin{aligned}
    2 \lambda_{\min}(\hat \K)^{-1} - \lambda_{\max}(\hat \K)^{-2} 
    + 2((\lambda_{\max}(\K) / \lambda_{\min}(\K) - 1) \lambda_{\min}(\hat \K)^{-2} \lambda_{\min}(\K)^{-1} \|\y\|_2^2 < M. \\  
\end{aligned}
\]
Re-arranging, we attain
\begin{align}
    \lambda_{\min} \hat\lambda_{\min}^2  
    \frac{(M (1 - s_{\max})^{2} + \hat \lambda_{\min}^{-2} - 2 \hat \lambda_{\max}^{-1})}{2(\lambda_{\max} / \lambda_{\min} - 1)} > \| \y \|_2^2,
\end{align}
which is a non-trivial guarantee when 
$M (1 - s_{\max})^{2} + \hat \lambda_{\min}^{-2} - 2 \hat \lambda_{\max}^{-1} > 0$.
\end{proof}

\ddconcavity*
\begin{proof}
    Fist, Gershgorin's Disk Theorem implies that the eigenvalues of
    $\diag(\mathbf A)^{-1/2} \mathbf A \diag(\mathbf A)^{-1/2}$ lie in $(1 - \delta, 1 + \delta)$
    for a $\delta$-diagonally dominant matrix $\mathbf A$.
    Further, the condition number of $\mathbf A$ is bounded above by
    $\kappa(\mathbf A) = \lambda_{\max}(\mathbf A) / \lambda_{\min}(\mathbf A) \leq (1 + \delta) / (1 - \delta)$.
    See \citet{horn2012matrix} for more background on matrix analysis.
    Plugging these bounds into the results of Lemma~\ref{lem:Theory:Convexity:EvalCondition} yields
    \[
    \begin{aligned}
        \lambda_{\min}(\K) \hat\lambda_{\min}^2  
    \frac{(2 \hat \lambda_{\max}^{-1} - \hat \lambda_{\min}^{-2} - m)}{2(1 -  \lambda_{\min} / \lambda_{\max})} 
    &\geq \lambda_{\min}(\K) (1 - \delta)^2 
    \frac{(2 (1 + \delta)^{-1} - (1-\delta)^{-2} - m)}{2(1 -  (1 - \delta) / (1 + \delta))}.
    \end{aligned}
    \]
    Lower bounding the last expression by $\|\y\|_2^2$ implies $m$-strong convexity.
    This gives rise to a non-trivial guarantee whenever the numerator is 
    larger than zero.
    In particular,
    \[
    2 (1 + \delta)^{-1} - (1-\delta)^{-2} - m > 0.
    \]
    Expanding, noting that $0 < \delta < 1$ implies $\delta^3 < \delta^2$
    in order to reduce a power in the resulting expression,
    and collecting terms with like powers, we attain
    the following sufficient condition
    \[
    2\delta^2 + (m-5) \delta + (1 - m) > 0. 
    \]
    Note that for this condition to hold if $\delta = 0$, 
    we need to have $m < 1$.
    Fortunately, this is a quadratic in $\delta$ whose smallest positive root is
    \[
    \delta_{-} = \frac{1}{4}\left( (5 - m) -  \sqrt{(5 - m)^2 - 8(1-m)}\right).
    \]
    In particular, for $m=0$, this reduces to
    \[
    \delta_{-} = 
    \frac{1}{4}\left( 5 - \sqrt{17}\right)
    \approx 0.4384471871911697.
    \]
    Lastly, note that if $\K_0$ is $\delta$-diagonally dominant, then
    so is $\K_\bs = \K_0 + \D_\brho(\bs)$,
    since the robust variances add to the diagonal,
    making it more dominant.
    Therefore, the convexity guarantee holds for all $\brho(\bs)$,
    if it holds for the base covariance matrix $\K_0$.
    Note that $\lambda_{\min}(\K) \geq \sigma^2$.
\end{proof}

Similarly, we can prove a similar statement relating diagonal dominance with $M$-smoothness.

\begin{restatable}{lem}{ddsmoothness}[Smoothness via Diagonal Dominance]
\label{lem:smoothness_via_diagonal_dominance}
    Suppose $\K_0$ is a $\delta$-diagonally dominant covariance matrix and
    suppose we constrain $\|\bs\|_\infty \leq s_{\max} \leq 1 - \sqrt{1 / M}$.
    Then
    \[
    \lambda_{\min}(\K_0) 
    (1-\delta)^2  
    \frac{M (1 - s_{\max})^{2} - 1}{2((1+\delta) / (1-\delta) - 1)}
    \geq 
    \| \y \|^2_2,
    \]
    implies that the NLML is $m$-strongly convex for all $\bs \in [0,1]^n$, i.e. $\brho(\mathbf s) \in [0, \infty]^n$.
\end{restatable}
\begin{proof}
    We proceed in a similar way as for Lemma~\ref{lem:strong_convexity_via_diagonal_dominance},
    but with Lemma~\ref{lem:Theory:Smoothness:EvalCondition} 
    as the starting point.
    \[
    \begin{aligned}
    \lambda_{\min}(\K) (\hat\lambda_{\min})^2  
    \frac{(M (1 - s_{\max})^{2} + \hat \lambda_{\min}^{-2} - 2 \hat \lambda_{\max}^{-1})}{2(\lambda_{\max} / \lambda_{\min} - 1)} 
    &\geq \lambda_{\min}(\K) 
    (1-\delta)^2  
    \frac{M (1 - s_{\max})^{2} - 1}{2((1+\delta) / (1-\delta) - 1)},
    \end{aligned}
    \]
    where we used 
    $(\hat \lambda_{\min}^{-2} - 2 \hat \lambda_{\max}^{-1}) \geq -1$,
    which is tight when $\hat \K$ is unitary.
    Lower bounding the last expression by $\|\y\|_2^2$ implies $M$-smoothness.
    This gives rise to a non-trivial guarantee whenever the numerator is 
    larger than zero.
    In particular,
    $M (1 - s_{\max})^{2} - 1 > 0$,
    which implies $s_{\max} \leq 1 - \sqrt{1 / M}$
    or equivalently,
    $M > 1 / (1 - s_{\max})^2$.
    Lastly, note that if $\K_0$ is $\delta$-diagonally dominant, then
    so is $\K_\bs = \K_0 + \D_\brho(\bs)$, since the robust variances only add to the diagonal.
    Therefore, if the inequality holds for the base covariance matrix $\K_0$,
    the smoothness guarantee holds for all $\brho(\bs)$
    such that $\bs \leq s_{\max} \leq 1 - \sqrt{1 / M}$.
    Note also that $\lambda_{\min}(\K) \geq \sigma^2$.
\end{proof}

\approxguarantee*
\begin{proof}
    The result is a direct consequence of meeting the $m$-convexity and $M$-smoothness conditions of Lemmas~\ref{lem:strong_convexity_via_diagonal_dominance}, \ref{lem:smoothness_via_diagonal_dominance} above,
    and the OMP approximation guarantee
    of Theorem~\ref{thm:omp_approximation_guarantee} due to \citet{elenberg2018restricted}.
    Note that the condition on $2r \leq n$ comes from 
    the RSC condition in Theorem~\ref{thm:omp_approximation_guarantee}
    being required for subsets of size $2r$.
    As we proved bounds for the $m$-convexity of the full Hessian of size $n$, $r$ has to be smaller than $n / 2$ for the assumptions of the theorem to hold.
    Regarding the upper bound $s_{\max}$ on $\bs$, we note that the constraint is convex and therefore doesn't change the convexity property of the optimization problem.
    
    Further, note that $\max_{\brho_{\support}} \calL(\brho_{\support}) \leq 
    \max_{\brho_{\support \cup i}} \calL(\brho_{\support \cup i})$,
    since the additional non-zero element could stay at $0$,
    if the marginal likelihood does not improve with $\rho_i$ increasing.
    That is, the subset selection problem is monotonic.
    As a consequence, we can normalize the MLL by $\tilde \calL(\cdot) = \calL(\cdot) - \calL(\mathbf 0)$, which then only takes positive values
for any $\bs^*_\support
= \arg \max_{\bs_{\backslash \support} = 0}
\calL(\brho(\bs_\support))$,
i.e. $\max_{\bs_\support} \tilde \calL(\bs_{\support}) \geq 0$.
This normalization is required for the constant factor approximation
guarantee to apply, similar to the original work of Nemhauser.
\end{proof}

This theoretical approach could lead to approximation guarantees for \citet{tipping2001sparse}'s Sparse Bayesian Learning (SBL) model,
for which 
\citet{ament2021sblviastepwise} show that greedily optimizing the associated NMLL
is equivalent to stepwise regression in the limit of $\sigma \to 0$,
proving exact recovery guarantees.

\section{Additional Detail on the Experiments}
\label{appdx:AddEmpirical}

Our benchmark uses a modified version of the code from \citet{andrade2023trimmed}, available at \url{https://github.com/andrade-stats/TrimmedMarginalLikelihoodGP} under the GNU GPLv2 license.

\subsection{Synthetic Regression Problems}
\label{appdx:AddEmpirical:Synthetic}

\begin{figure}[!ht]
    \centering
    \includegraphics[width=\textwidth]{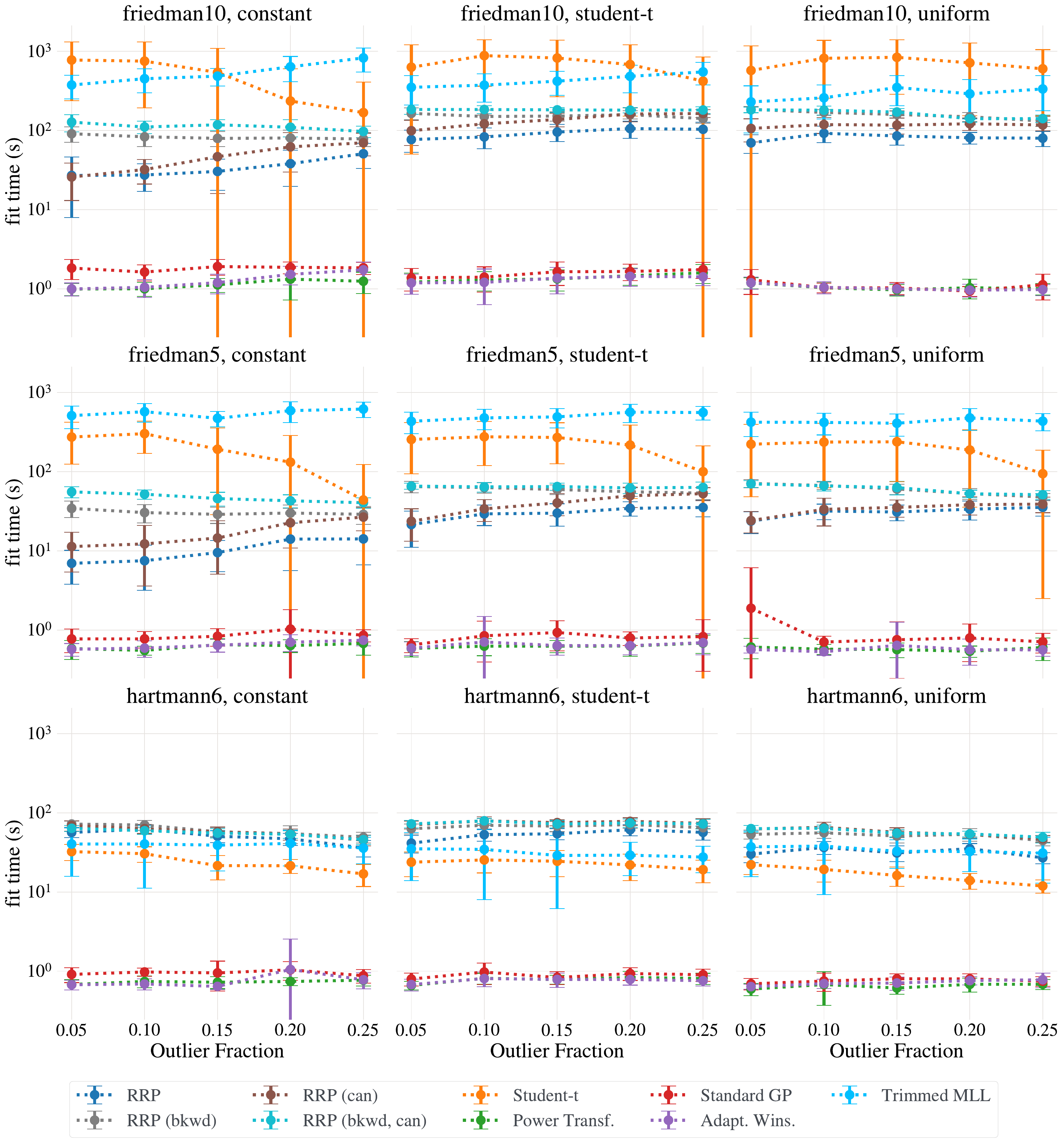}
    \caption{
    Fitting times of the different robust GP modeling approaches. The plots show means and one standard deviation. Here ``bkwd'' indicates the backward variant of RRP (Algorithm~\ref{algo:RobustRegressionModel:Inference:RelevancePursuitBackward}), and ``can'' indicates the canonical (non-convex) parameterization.}
    \label{fig::AddEmpirical:Synthetic:FitTimeComplete}
\end{figure}

\begin{figure}[!ht]
    \centering
    \includegraphics[width=\textwidth]{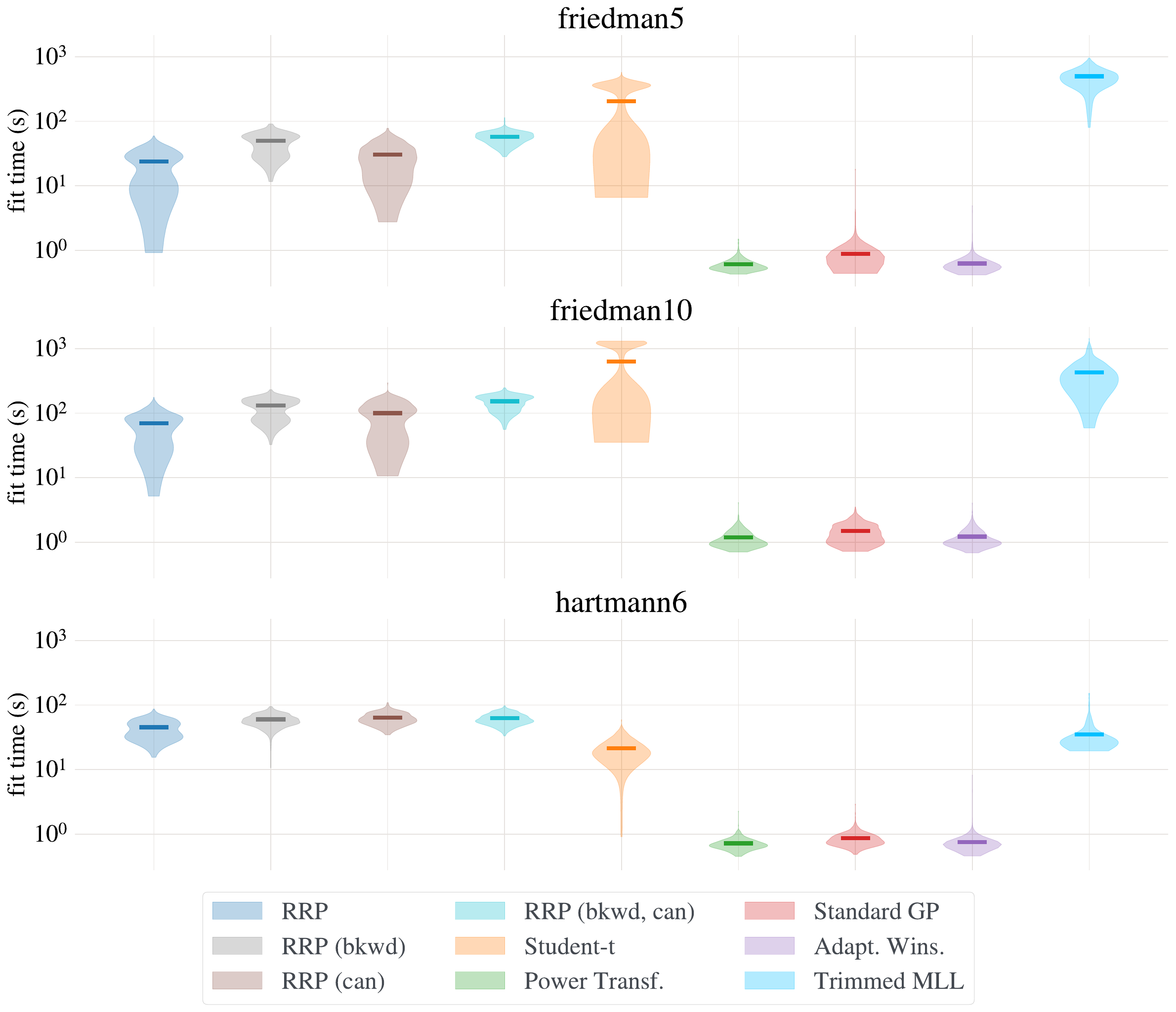}
    \caption{
    Fitting times of the different robust GP modeling approaches on the regression tasks from Section~\ref{subsubsec:EmpiricalResults:Regression:Synthetic}. Results are aggregated across outlier types and outlier fractions as those do not affect fitting times much (see Fig.~\ref{fig::AddEmpirical:Synthetic:FitTimeComplete}).}
    \label{fig:AddEmpirical:Synthetic:FitTimeAggregated}
\end{figure}

\paragraph{Model fitting runtimes} 
Fig.~\ref{fig::AddEmpirical:Synthetic:FitTimeComplete} summarizes the fitting times for the different models on the different scenarios from Section~\ref{subsubsec:EmpiricalResults:Regression:Synthetic}. 
We observe that the outlier type and the fraction of outliers both have relatively limited effect on the fitting times for all of the models.  
Fig.~\ref{fig:AddEmpirical:Synthetic:FitTimeAggregated} therefore provides a more compact view of the same data (aggregated across outlier types and outlier fractions), giving a better sense of the distribution of the fitting times. Unsurprisingly, the baselines that simply fit a single GP model (``vanilla'', ``winsorize'', ``power\_transform'') are substantially faster than any of the robust approaches. While all of the robust models show similar fitting times on the Hartmann problem, fitting our \ouralgo{} is significantly faster (note the logarithmic scale of the y-axis) than fitting the Student-$t$ and trimmed MLE models on the 5-dim and 10-dim Friedman problems.
The trimmed MLE model in particular ends up being quite slow, especially on the 5d Friedman function. 

\subsection{BO experiments, additional details}
\label{appdx:AddEmpirical:bo_details}
For Hartmann6, we consider the standard domain of $[0, 1]^6$.

\paragraph{SVM}
The SVM problem, the goal is to optimize the test RMSE of an SVM regression model trained on $100$ features from the CT slice UCI dataset.
We tune the following three parameters: $C \in$ [1e-2, 1e2], $\epsilon \in$ [1e-2, 1], and $\gamma \in$ [1e-3, 1e-1].
All parameters are tuned in log-scale.
Corruptions simulate I/O failures in which case we only train on $U[100, 1000]$ training points out of the available $50,000$ training observations.

\paragraph{CNN}
For the 5D CNN problem the goal is to optimize the test accuracy of a CNN classifier trained on the MNIST dataset. 
We tune the following 5 parameters: learning rate in the interval [1e-4, 1e-1], momentum in the interval [0, 1], weight decay in the interval [0, 1], step size in the interval [1, 100], and $\gamma \in$ [0, 1].
Similarly to the SVM problem, we only train on $U[100, 1000]$ of the available training batches when an I/O failure occurs.

\paragraph{Rover}
The rover trajectory planning problem was originally proposed in~\citet{ebo}.
The goal is to tune the trajectory of a rover in order to maximize the reward of its final trajectory.
We use the same obstacle locations and trajectory parameterization as in~\citet{robot}, with the main difference being that we parameterize the trajectory using $10$ points, resulting in $20$ tunable parameters.
When a corruption occurs, the rover will stop at a uniformly random point along its trajectory, generally resulting in lower reward than the original trajectory.

\subsection{Additional Synthetic BO Results}
\label{appdx:additional_bo}
\begin{figure}[!ht]
\centering
\includegraphics[width=0.98\textwidth]{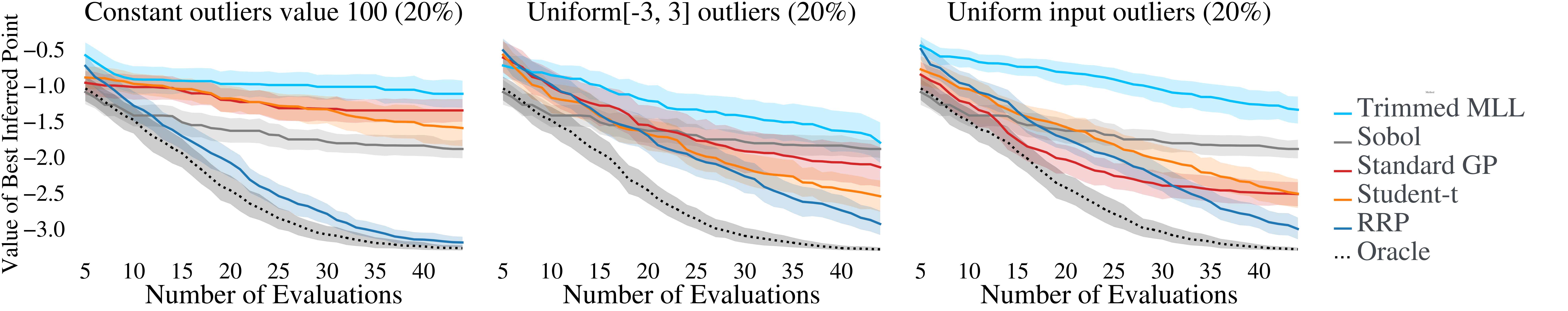}
\caption{BO results for Hartmann6: \textit{Left:} We see that Relevance pursuit performs well in the case of constant outliers of value $100$ and almost performs just as well as the oracle. \textit{Middle:} Relevance pursuit performs the best followed by the Student-$t$ likelihood from \cite{martinezcantin2017robust} in the case of $U[-3, 3]$. No method performs as well as the oracle when the outlier probability is 20\%. \textit{Right:} Similarly to the middle column, this setting hides the outliers within the range of the outliers making it difficult to match the performance of the oracle.}
\label{fig:bo_hartmann_2}
\end{figure}

In addition to the results in Fig.~\ref{fig:bo_hartmann}, we also include results when the corruption probability is 20\% in Fig.~\ref{fig:bo_hartmann_2}. 
We observe that the results are similar as with 10\% corruption probability, but that the performance of several baselines regresses significantly.

\subsection{Computational Setup and Requirements}
\label{appdx:AddEmpirical:Compute}

Robust GP regression is very data-efficient, focuses on the small-data regime, and runs fast (faster than competing baselines studied in this paper). 
Therefore, each individual run required very limited compute resources (this includes the baseline methods). To produce statistically meaningful results, however, we ran a large number of replications for both our regression and Bayesian optimization benchmarks on a proprietary cluster. We estimate the amount of compute spent on these experiments to be around 2 CPU years in total, using standard (Intel Xeon) CPU hardware. The amount of compute spent on exploratory investigations as part of this work was negligible (this was ad-hoc exploratory and development work on a single CPU machine).

\subsection{Impact of Convex Parameterization on Joint Optimization of Hyper-Parameters}
\label{subsec:convex_parameterization_for_joint_optimization}

A limitation of Lemma~\ref{lem:Theory:Convexity:ReparamHessian} is that it only guarantees convexity for the sub-problem of optimizing the $\rho_i$'s.
In practice, we jointly optimize all GP hyper-parameters, including length scales.
In this case, the theory guarantees the positive-definiteness of the {\it submatrix} of the Hessian corresponding to $\brho(\bs)$,
and we expect this to improve the quality of the results of the numerical optimization routines.

Indeed, positive-definiteness is beneficial for quasi-Newton optimization algorithms like L-BFGS~\cite{nocedal1989lbfgs}, which restarts its approximation to the Hessian whenever it encounters non-convex regions, because the associated updates to the Hessian approximation are not positive-definite. This leads the algorithm to momentarily revert back to gradient descent, with an associated slower convergence rate.

To quantify the impact of this, we ran convergence analyses using the data from Fig.~\ref{fig:sine_example}, allowing all $\brho$ to be optimized jointly with other hyper-parameters (length scale, kernel variance and noise variance), recording the achieved negative log marginal likelihood (NLML) as a function of the tolerance parameter ftol of the L-BFGS optimizer. The results are reported in Table~\ref{table:convergence_comparison}, and
indicate that the optimizer terminates with a much better NLML using the convex parameterization with the same convergence tolerance.

\begin{table}
\begin{minipage}[c]{0.49\textwidth}
\begin{center}
\begin{tabular}{ ccc } 
 \toprule
 \texttt{ftol} & Canonical $\brho$ & Convex $\brho(\bs)$ \\ 
 \midrule
    $1\mathrm{e}{-3}$ &	$-4.37$ & $\mathbf{-14.18}$ \\
    $1\mathrm{e}{-4}$ &	$-4.37$ & $\mathbf{-93.00}$ \\
    $1\mathrm{e}{-5}$ &	$-4.37$ & $\mathbf{-93.01}$ \\
    $1\mathrm{e}{-6}$ &	$-4.37$ & $\mathbf{-135.05}$ \\
    $1\mathrm{e}{-7}$ &	$-98.62$ & $\mathbf{-518.68}$ \\
    $1\mathrm{e}{-8}$ &	$-97.52$ & $\mathbf{-1139.29}$ \\
 \bottomrule
\end{tabular}
\end{center}
    \end{minipage}\hfill
    \begin{minipage}[c]{0.45\textwidth}
        \caption{
        Comparison of the negative log marginal likelihood achieved after numerical optimization of the canonical and convex parameterization of $\brho$ with L-BFGS.
        Notably, the convex parameterization leads to improved marginal likelihood values for for a given convergence tolerance \texttt{ftol}. 
        }
    \end{minipage}
\label{table:convergence_comparison}
\end{table}

There are also settings in which we do not actually jointly optimize the hyper-parameters, particularly when we have access to data from the same data generating process that has been manually labeled by domain experts as outlier-free. 
Then we can estimate the model hyper-parameters on that data, and fix them for the \ouralgo{} on new data sets that we do not know to be outlier-free.

\subsection{Comparison to Robust and Conjugate Gaussian Processes (RCGP)}
\label{appdx:AddEmpirical:rcgp}

We report extended empirical comparisons with \citet{altamirano2023robust}'s RCGP method, using their 
experimental setup and method
\href{https://github.com/maltamiranomontero/RCGP}{implementation in GPFLow}.
Including GPFlow in our own benchmarking setup and compute resources proved difficult.
To circumvent this, we wrote wrappers for both BoTorch's standard GP and \ouralgo{}, which also accounts for any orthogonal implementation differences between the two frameworks, and ran the benchmarks locally on an M-series MacBook.

See Tables~\ref{tab:rcgp_comparison:mae} and \ref{tab:rcgp_comparison:nlpd} 
for the mean absolute error and negative log predictive density, respectively.
The tables include the empirical mean and standard deviation over 20 replications
on corrupted version of the following base data sets:
1) Synthetic, which is generated as a draw of a GP with a exponentiated quadratic kernel,
and four data sets available on the UCI machine learning repository\citep{kelly2023uci}, in particular,
2) Boston~\citep{HarrisonRubinfeld1978boston_housing},
3) Concrete~\citep{concrete_compressive_strength_165},
4) Energy~\citep{energy_efficiency_242}, and
5) Yacht~\citep{yacht_hydrodynamics_243}.
The benchmark considers no corruptions (``No Outliers''),
``Asymmetric Outliers'',
which are uniform in $x$ are shifted negatively in $y$,
``Uniform Outliers'',
which shift $y$ in both directions (positively and negatively),
and 
``Focused Outliers'',
which form concentrated clusters in both $x$ and $y$.
Any bold entry in the table signifies a results that is
within one standard-error of the best result's one standard-error confidence bound.

\begin{table}[h]
    \centering
    \caption{Mean absolute error (MAE) using \citet{altamirano2023robust}'s experimental setup in GPFlow.
    \ouralgo{} is always competitive with the other methods,
    and outperforms them significantly for uniform and asymmetric outliers.
    }
    \label{tab:rcgp_comparison:mae}
    \tiny
    \begin{tabular}{lccccccc}
    \toprule
    & Standard GP (GPFlow) & Student-$t$ GP (GPFLow) & RCGP (GPFlow) & Standard GP (BoTorch) & \ouralgo{} (BoTorch) \\
    \midrule
    No Outliers \\
    Synthetic &  \bf 8.82e-2 (2.12e-3) &  \bf 8.81e-2 (2.09e-3) &  \bf 8.77e-2 (2.08e-3) &  \bf 8.90e-2 (2.19e-3) &  \bf 8.90e-2 (2.19e-3) \\
    Boston & 2.24e-1 (7.11e-3) & 2.24e-1 (6.50e-3) & 2.24e-1 (7.29e-3) &  \bf 2.08e-1 (5.53e-3) &  \bf 2.08e-1 (5.75e-3) \\
    Concrete & 2.00e-1 (3.72e-3) &  \bf 1.99e-1 (3.73e-3) &  \bf 1.93e-1 (4.05e-3) &  \bf 1.92e-1 (3.80e-3) &  \bf 1.93e-1 (3.66e-3) \\
    Energy & 3.01e-2 (8.65e-4) & 3.44e-2 (1.46e-3) &  \bf 2.42e-2 (1.01e-3) & 3.09e-2 (9.41e-4) & 3.03e-2 (8.88e-4) \\
    Yacht &  \bf 1.69e-2 (2.09e-3) & 2.02e-2 (1.29e-3) & 2.19e-2 (4.63e-3) &  \bf 1.49e-2 (1.95e-3) &  \bf 1.42e-2 (1.58e-3) \\
    \midrule
    Uniform Outliers \\
    Synthetic & 3.45e-1 (1.48e-2) & 2.93e-1 (1.23e-2) & 2.15e-1 (9.30e-3) & 3.48e-1 (1.62e-2) &  \bf 8.99e-2 (2.37e-3) \\
    Boston & 4.97e-1 (2.80e-2) & 3.86e-1 (1.52e-2) & 4.94e-1 (1.81e-2) & 6.81e-1 (2.25e-2) &  \bf 2.83e-1 (3.16e-2) \\
    Concrete & 3.80e-1 (8.36e-3) & 3.54e-1 (5.98e-3) & 4.43e-1 (7.97e-3) & 3.80e-1 (7.77e-3) &  \bf 2.03e-1 (3.98e-3) \\
    Energy & 2.67e-1 (8.93e-3) & 2.73e-1 (2.40e-2) & 3.05e-1 (3.55e-2) & 5.60e-1 (1.47e-2) &  \bf 3.06e-2 (8.35e-4) \\
    Yacht & 3.23e-1 (1.68e-2) & 2.29e-1 (1.54e-2) & 4.18e-1 (3.48e-2) & 6.81e-1 (3.21e-2) &  \bf 1.20e-2 (1.04e-3) \\
    \midrule
    Asymmetric Outliers \\
    Synthetic & 1.11e+0 (2.24e-2) & 1.02e+0 (2.34e-2) & 7.72e-1 (3.31e-2) & 1.17e+0 (2.55e-2) &  \bf 2.71e-1 (9.96e-2) \\
    Boston & 6.92e-1 (1.40e-2) & 5.80e-1 (1.24e-2) & 5.31e-1 (2.33e-2) & 8.55e-1 (3.04e-2) &  \bf 3.35e-1 (4.33e-2) \\
    Concrete & 6.60e-1 (1.21e-2) & 5.55e-1 (1.08e-2) & 4.91e-1 (1.11e-2) & 6.60e-1 (1.25e-2) &  \bf 2.03e-1 (4.45e-3) \\
    Energy & 6.03e-1 (1.12e-2) & 4.74e-1 (9.99e-3) & 4.08e-1 (3.34e-2) & 7.05e-1 (1.78e-2) &  \bf 3.07e-2 (8.33e-4) \\
    Yacht & 5.90e-1 (1.60e-2) & 4.58e-1 (9.31e-3) & 4.32e-1 (3.30e-2) & 8.36e-1 (3.76e-2) &  \bf 2.70e-2 (1.35e-2) \\
    \midrule
    Focused Outliers \\
    Synthetic &  \bf 1.89e-1 (1.50e-2) &  \bf 1.92e-1 (1.56e-2) &  \bf 1.64e-1 (1.29e-2) &  \bf 1.80e-1 (1.46e-2) & 2.00e-1 (1.62e-2) \\
    Boston &  \bf 2.44e-1 (8.48e-3) &  \bf 2.60e-1 (1.30e-2) &  \bf 2.49e-1 (1.04e-2) &  \bf 2.49e-1 (8.37e-3) &  \bf 2.48e-1 (8.38e-3) \\
    Concrete & 2.40e-1 (5.54e-3) &  \bf 2.35e-1 (5.63e-3) &  \bf 2.38e-1 (1.10e-2) &  \bf 2.24e-1 (6.20e-3) &  \bf 2.25e-1 (5.86e-3) \\
    Energy & 8.83e-2 (5.46e-2) & 3.24e-2 (1.36e-3) &  \bf 2.92e-2 (8.95e-4) &  \bf 3.03e-2 (8.95e-4) &  \bf 3.03e-2 (8.95e-4) \\
    Yacht & 2.42e-1 (6.89e-2) & 1.10e-1 (2.07e-2) &  \bf 1.93e-2 (1.88e-3) &  \bf 1.78e-2 (2.96e-3) &  \bf 1.59e-2 (2.20e-3) \\
    \bottomrule
    \end{tabular}
\end{table}

\begin{table}[h!]
    \centering
    \caption{Negative log predictive density (NLPD) using \citet{altamirano2023robust}'s experimental setup in GPFlow.
    \ouralgo{} is generally competitive,
    and outperforms other methods significantly for uniform and asymmetric outliers.
    }
    \label{tab:rcgp_comparison:nlpd}
    \tiny
    \begin{tabular}{lccccccc}
    \toprule
    & Standard GP (GPFlow) & Student-$t$ GP (GPFLow) & RCGP (GPFlow) & Standard GP (BoTorch) & \ouralgo{} (BoTorch) \\
    \midrule
    No Outliers \\
    Synthetic &  \bf -8.21e-1 (2.16e-2) &  \bf -8.20e-1 (2.16e-2) &  \bf -8.23e-1 (2.21e-2) &  \bf -8.05e-1 (2.25e-2) &  \bf -8.05e-1 (2.24e-2) \\
    Boston & 2.08e-1 (4.26e-2) & 1.99e-1 (3.76e-2) & 1.95e-1 (4.69e-2) &  \bf 9.24e-2 (2.82e-2) &  \bf 9.43e-2 (2.82e-2) \\
    Concrete &  \bf 1.48e-1 (2.33e-2) &  \bf 1.29e-1 (2.31e-2) &  \bf 1.09e-1 (3.19e-2) &  \bf 1.11e-1 (2.73e-2) &  \bf 1.16e-1 (2.68e-2) \\
    Energy & -1.72e+0 (3.83e-2) & -1.62e+0 (3.46e-2) &  \bf -1.96e+0 (4.37e-2) & -1.67e+0 (3.08e-2) & -1.69e+0 (4.02e-2) \\
    Yacht & -1.79e+0 (3.23e-1) & -2.05e+0 (6.61e-2) & -2.00e+0 (3.27e-1) &  \bf -2.47e+0 (1.06e-1) &  \bf -2.23e+0 (1.84e-1) \\
    \midrule
    Uniform Outliers \\
    Synthetic & 1.60e+0 (1.41e-2) & 1.47e+0 (1.55e-2) & 1.57e+0 (1.39e-2) & 1.58e+0 (1.40e-2) &  \bf -7.99e-1 (2.26e-2) \\
    Boston & 1.67e+0 (7.55e-3) & 1.54e+0 (8.98e-3) & 1.64e+0 (1.12e-2) & 1.81e+0 (5.95e-2) &  \bf 4.52e-1 (1.34e-1) \\
    Concrete & 1.64e+0 (6.83e-3) & 1.51e+0 (6.56e-3) & 1.63e+0 (7.57e-3) & 1.63e+0 (6.72e-3) &  \bf 1.47e-1 (2.51e-2) \\
    Energy & 1.61e+0 (7.75e-3) & 1.50e+0 (1.76e-2) & 1.61e+0 (9.74e-3) & 1.70e+0 (8.58e-3) &  \bf -1.67e+0 (3.72e-2) \\
    Yacht & 1.61e+0 (9.50e-3) & 1.46e+0 (1.19e-2) & 1.62e+0 (1.87e-2) & 1.78e+0 (4.67e-2) &  \bf -2.32e+0 (1.97e-1) \\
    \midrule
    Asymmetric Outliers \\
    Synthetic & 1.94e+0 (9.39e-3) & 1.90e+0 (9.98e-3) & 1.89e+0 (1.13e-2) & 1.93e+0 (1.06e-2) &  \bf -3.70e-1 (2.30e-1) \\
    Boston & 1.68e+0 (1.01e-2) & 1.56e+0 (1.09e-2) & 1.64e+0 (9.51e-3) & 2.52e+0 (7.26e-1) &  \bf 6.14e-1 (1.19e-1) \\
    Concrete & 1.66e+0 (7.43e-3) & 1.54e+0 (7.45e-3) & 1.62e+0 (6.63e-3) & 1.65e+0 (7.55e-3) &  \bf 1.55e-1 (2.86e-2) \\
    Energy & 1.62e+0 (8.85e-3) & 1.49e+0 (9.70e-3) & 1.62e+0 (1.82e-2) & 1.71e+0 (9.79e-3) &  \bf -1.66e+0 (3.55e-2) \\
    Yacht &  \bf 1.59e+0 (9.35e-3) &  \bf 1.45e+0 (1.01e-2) &  \bf 1.56e+0 (8.15e-3) & 1.75e+0 (1.41e-2) &  \bf -2.17e-1 (1.82e+0) \\
    \midrule
    Focused Outliers \\
    Synthetic & 6.78e-1 (6.59e-2) &  \bf 6.11e-1 (8.68e-2) &  \bf 5.69e-1 (4.07e-2) &  \bf 6.66e-1 (6.60e-2) & 9.67e+0 (2.24e+0) \\
    Boston &  \bf 2.57e-1 (4.23e-2) & 3.21e-1 (6.40e-2) &  \bf 2.74e-1 (6.13e-2) &  \bf 1.98e-1 (3.04e-2) &  \bf 1.97e-1 (3.06e-2) \\
    Concrete & 2.77e-1 (2.58e-2) &  \bf 2.72e-1 (2.81e-2) &  \bf 2.43e-1 (4.26e-2) &  \bf 2.22e-1 (2.87e-2) &  \bf 2.28e-1 (2.72e-2) \\
    Energy & 2.81e+4 (2.81e+4) &  \bf -1.67e+0 (4.13e-2) &  \bf -1.74e+0 (4.98e-2) &  \bf -1.70e+0 (4.59e-2) &  \bf -1.70e+0 (4.59e-2) \\
    Yacht & 9.38e+4 (5.09e+4) & -4.60e-1 (2.91e-1) & -2.47e+0 (8.09e-2) &  \bf -2.63e+0 (5.55e-2) &  \bf -2.61e+0 (5.66e-2) \\
    \bottomrule
    \end{tabular}
\end{table}

\subsection{Comparison to Robust Gaussian Process with Huber Likelihood}

In the following, we compare our method
with additional variational GP baselines with Laplace and Huber likelihoods, and translated the Matlab code of the "projection statistics" of \citet{algikar2023robustgaussianprocessregression} to PyTorch. We then combined the projection-statistics-based weighting of the Huber loss with a variational (referred to as Huber-Projection) to get as close as possible to a direct comparison to \citet{algikar2023robustgaussianprocessregression} without access to a Matlab license.

Tables \ref{table:rmse_15_percent_huber}
and \ref{table:nlpd_15_percent_huber}
shows the root mean square error and negative log predictive density on the Neal, Friedman 5 and Friedman 10 test functions,
as well as the Yacht Hydrodynamics~\citep{yacht_hydrodynamics_243}
and California Housing~\citep{pace1997cahousing} 
datasets from the UCI database~\citep{kelly2023uci}, where 15\% of the training data sets of the models were corrupted.
Tables \ref{table:rmse_100_percent_huber}
and 
\ref{table:nlpd_100_percent_huber} below 
were generated in a similar way, but  
100\% of the data were subject to heavier-tailed noise, either Student-$t$ or Laplace.

In summary, the Relevance Pursuit model generally outperforms the variational GPs with heavy-tailed likelihoods when the corruptions are a sparse subset of all observations.
Unsurprisingly, the GPs with heavy-tailed likelihoods perform best when {\it all} observations are subject to heavy-tailed noise.
While such uniformly heavy-tailed noise does exist in practice, 
we stress that this is a distinct setting to 
the common setting where datasets contain a {\it subset} of a-priori unknown outliers, while a dominant fraction of the data can be considered inliers that, once they are identified, can be used to train a model without additional treatment. 

\begin{table}[h]
\caption{Comparison with Huber GP: Root mean square error with 15\% Corruptions}
\label{table:rmse_15_percent_huber}
\tiny
\centering
\begin{tabular}{lccccccc}
\toprule
Data & Standard & Relevance Pursuit & Student-$t$ & Laplace & Huber & Huber + Projection \\
\midrule
Neal \\
\qquad Uniform & 3.87e-1 (2.40e-2) & \textbf{3.79e-2 (1.10e-2)} & 4.18e-1 (8.65e-3) & 4.18e-1 (4.72e-3) & 4.18e-1 (4.76e-3) & 4.18e-1 (4.76e-3) \\
\qquad Constant & 1.91e+0 (1.18e-1) & \textbf{4.37e-2 (1.87e-2)} & 7.26e-1 (1.38e-1) & 4.73e-1 (1.28e-2) & 4.89e-1 (1.62e-2) & 4.89e-1 (1.62e-2) \\
\qquad Student-$t$ & 7.74e-1 (2.70e-1) & \textbf{6.71e-2 (2.29e-2)} & 3.90e-1 (3.59e-3) & 4.07e-1 (4.79e-3) & 4.09e-1 (5.02e-3) & 4.09e-1 (5.02e-3) \\
\qquad Laplace & 7.19e-1 (6.17e-2) & \textbf{6.27e-2 (1.52e-2)} & 4.01e-1 (4.42e-3) & 4.20e-1 (4.34e-3) & 4.20e-1 (4.00e-3) & 4.20e-1 (4.00e-3) \\
\midrule
Friedman 5 \\
\qquad Uniform & 2.05e+0 (8.83e-2) & \textbf{9.40e-2 (1.23e-2)} & 7.01e-1 (1.37e-1) & 7.30e-1 (7.33e-2) & 6.78e-1 (6.67e-2) & 7.12e-1 (7.50e-2) \\
\qquad Constant & 1.48e+1 (5.16e-1) & \textbf{7.29e-2 (5.40e-3)} & 8.98e-1 (1.40e-1) & 2.37e+0 (5.19e-2) & 2.50e+0 (7.18e-2) & 2.50e+0 (7.18e-2) \\
\qquad Student-$t$ & 7.91e+0 (8.80e-1) & \textbf{1.17e-1 (2.24e-2)} & 5.26e-1 (1.26e-1) & 1.27e+0 (9.49e-2) & 1.36e+0 (9.34e-2) & 1.31e+0 (9.73e-2) \\
\qquad Laplace & 7.59e+0 (4.98e-1) & \textbf{9.63e-2 (1.15e-2)} & 6.62e-1 (1.34e-1) & 1.51e+0 (7.78e-2) & 1.63e+0 (6.89e-2) & 1.63e+0 (6.89e-2) \\
\midrule
Friedman 10 \\
\qquad Uniform & 1.67e+0 (4.94e-2) & \textbf{4.84e-2 (6.21e-3)} & 5.92e-1 (1.36e-1) & 6.82e-1 (1.03e-1) & 6.74e-1 (1.04e-1) & 5.93e-1 (9.43e-2) \\
\qquad Constant & 1.34e+1 (2.97e-1) & \textbf{5.12e-2 (2.69e-3)} & 8.63e-1 (1.39e-1) & 2.10e+0 (2.39e-2) & 2.18e+0 (3.01e-2) & 2.18e+0 (3.01e-2) \\
\qquad Student-$t$ & 7.93e+0 (1.75e+0) & \textbf{8.45e-2 (2.42e-2)} & 5.30e-1 (1.30e-1) & 1.40e+0 (9.10e-2) & 1.40e+0 (9.12e-2) & 1.40e+0 (9.04e-2) \\
\qquad Laplace & 6.42e+0 (3.42e-1) & \textbf{5.70e-2 (8.16e-3)} & 7.23e-1 (1.39e-1) & 1.64e+0 (4.98e-2) & 1.72e+0 (1.06e-2) & 1.72e+0 (1.06e-2) \\
\midrule
Yacht \\
\qquad Uniform & 6.72e+0 (5.08e-1) & \textbf{8.56e-1 (4.99e-2)} & 1.62e+1 (1.81e-1) & 1.62e+1 (2.08e-1) & 1.61e+1 (2.04e-1) & 1.61e+1 (1.94e-1) \\
\qquad Constant & 3.50e+1 (1.61e+0) & \textbf{8.27e-1 (5.78e-2)} & 1.60e+1 (2.11e-1) & 1.59e+1 (2.02e-1) & 1.55e+1 (1.73e-1) & 1.58e+1 (1.95e-1) \\
\qquad Student-$t$ & 2.11e+1 (3.51e+0) & \textbf{1.12e+0 (2.03e-1)} & 1.64e+1 (2.17e-1) & 1.67e+1 (2.15e-1) & 1.64e+1 (2.08e-1) & 1.66e+1 (2.02e-1) \\
\qquad Laplace & 1.61e+1 (1.97e+0) & \textbf{1.15e+0 (1.84e-1)} & 1.63e+1 (2.10e-1) & 1.64e+1 (4.06e-1) & 1.63e+1 (2.05e-1) & 1.66e+1 (1.99e-1) \\
\midrule
CA Housing \\
\qquad Uniform & \textbf{7.10e-1 (7.58e-3)} & 7.39e-1 (1.75e-2) & 1.16e+0 (1.79e-3) & 1.17e+0 (3.04e-3) & 1.17e+0 (2.93e-3) & 1.18e+0 (6.17e-3) \\
\qquad Constant & 2.28e+0 (5.12e-2) & \textbf{6.35e-1 (4.38e-3)} & 1.17e+0 (2.39e-3) & 1.16e+0 (1.87e-3) & 1.16e+0 (1.88e-3) & 1.17e+0 (5.51e-3) \\
\qquad Student-$t$ & 1.34e+0 (1.68e-1) & \textbf{6.56e-1 (5.46e-3)} & 1.18e+0 (3.56e-3) & 1.19e+0 (4.13e-3) & 1.18e+0 (3.83e-3) & 1.19e+0 (6.74e-3) \\
\qquad Laplace & 1.00e+0 (5.04e-2) & \textbf{6.51e-1 (4.71e-3)} & 1.18e+0 (3.41e-3) & 1.18e+0 (3.91e-3) & 1.18e+0 (3.67e-3) & 1.18e+0 (6.30e-3) \\
\bottomrule
\end{tabular}
\end{table}

\begin{table}
\caption{Comparison with Huber GP: Negative log predictive density with 15\% Corruptions}
\label{table:nlpd_15_percent_huber}
\tiny
\begin{tabular}{lccccccc}
\toprule
Data & Standard & Relevance Pursuit & Student-$t$ & Laplace & Huber & Huber + Projection \\
\midrule
Neal \\
\qquad Uniform & 6.87e+0 (4.15e+0) & \textbf{3.80e+0 (4.01e+0)} & \textbf{1.85e+0 (9.93e-2)} & \textbf{1.64e+0 (1.20e-1)} & \textbf{1.70e+0 (1.23e-1)} & \textbf{1.70e+0 (1.23e-1)} \\
\qquad  Constant & 1.32e+2 (1.29e+2) & \textbf{-2.50e+0 (2.99e-1)} & 2.03e+0 (1.84e-1) & 8.67e-1 (4.32e-2) & 8.84e-1 (3.87e-2) & 8.84e-1 (3.87e-2) \\
\qquad  Student-$t$ & \textbf{1.98e+0 (9.67e-1)} & \textbf{1.99e-1 (2.29e+0)} & \textbf{1.67e+0 (1.17e-1)} & \textbf{1.25e+0 (8.32e-2)} & \textbf{1.28e+0 (8.03e-2)} & \textbf{1.28e+0 (8.03e-2)} \\
\qquad  Laplace & 3.07e+2 (2.15e+2) & \textbf{2.02e+0 (2.83e+0)} & 1.73e+0 (1.10e-1) & \textbf{1.20e+0 (8.89e-2)} & \textbf{1.21e+0 (8.99e-2)} & \textbf{1.21e+0 (8.99e-2)} \\
\midrule
Friedman 5 \\
\qquad Uniform & 2.07e+0 (4.88e-2) & \textbf{-1.35e+0 (7.22e-2)} & 4.67e-1 (3.00e-1) & 8.88e-1 (1.01e-1) & 8.29e-1 (9.14e-2) & 8.88e-1 (1.08e-1) \\
\qquad Constant & 4.28e+0 (2.25e-2) & \textbf{-1.13e+0 (1.82e-2)} & 8.24e-1 (3.17e-1) & 2.28e+0 (2.63e-2) & 2.34e+0 (3.13e-2) & 2.34e+0 (3.13e-2) \\
\qquad Student-$t$ & 3.42e+0 (1.34e-1) & \textbf{-1.11e+0 (1.04e-1)} & 1.78e-2 (2.86e-1) & 1.55e+0 (1.14e-1) & 1.66e+0 (1.11e-1) & 1.59e+0 (1.10e-1) \\
\qquad Laplace & 3.55e+0 (7.84e-2) & \textbf{-1.22e+0 (5.89e-2)} & 3.47e-1 (3.13e-1) & 1.75e+0 (8.46e-2) & 1.90e+0 (7.36e-2) & 1.90e+0 (7.36e-2) \\
\midrule
Friedman 10 \\
\qquad Uniform & 1.86e+0 (3.04e-2) & \textbf{-1.82e+0 (3.97e-2)} & 6.78e-2 (3.85e-1) & 9.20e-1 (1.99e-1) & 9.39e-1 (2.01e-1) & 7.65e-1 (1.80e-1) \\
\qquad Constant & 4.21e+0 (1.39e-2) & \textbf{-1.27e+0 (1.10e-2)} & 8.85e-1 (3.79e-1) & 2.20e+0 (1.45e-2) & 2.23e+0 (1.54e-2) & 2.23e+0 (1.54e-2) \\
\qquad Student-$t$ & 3.34e+0 (1.24e-1) & \textbf{-1.40e+0 (1.17e-1)} & -6.80e-2 (3.65e-1) & 1.96e+0 (1.25e-1) & 1.95e+0 (1.23e-1) & 1.96e+0 (1.22e-1) \\
\qquad Laplace & 3.46e+0 (6.68e-2) & \textbf{-1.44e+0 (5.91e-2)} & 4.89e-1 (3.94e-1) & 2.13e+0 (6.35e-2) & 2.21e+0 (2.01e-2) & 2.21e+0 (2.01e-2) \\
\midrule
Yacht \\
\qquad Uniform & 3.45e+0 (1.42e-1) & \textbf{2.37e+0 (2.79e-1)} & 1.18e+2 (1.06e+1) & 7.07e+1 (5.17e+0) & 7.46e+1 (5.38e+0) & 1.89e+2 (1.64e+1) \\
\qquad Constant & 5.29e+0 (1.70e-1) & \textbf{1.79e+0 (2.43e-1)} & 7.16e+1 (7.68e+0) & 2.32e+1 (1.59e+0) & 1.95e+1 (1.12e+0) & 3.89e+1 (3.46e+0) \\
\qquad Student-$t$ & 4.46e+0 (3.08e-1) & \textbf{2.42e+0 (3.38e-1)} & 1.32e+2 (1.05e+1) & 8.20e+1 (5.07e+0) & 8.13e+1 (5.17e+0) & 1.55e+2 (1.38e+1) \\
\qquad Laplace & 4.35e+0 (2.88e-1) & \textbf{2.27e+0 (3.71e-1)} & 1.14e+2 (9.86e+0) & 6.85e+1 (4.72e+0) & 6.12e+1 (3.87e+0) & 1.17e+2 (8.89e+0) \\
\midrule
CA Housing \\
\qquad Uniform & \textbf{3.14e+0 (1.44e-1)} & \textbf{2.92e+0 (2.06e-1)} & 5.85e+1 (9.59e-1) & 6.77e+1 (2.41e+0) & 6.69e+1 (1.94e+0) & 8.94e+1 (2.45e+1) \\
\qquad Constant & \textbf{3.57e+0 (2.26e-1)} & 4.51e+0 (2.11e-1) & 4.02e+1 (1.59e+0) & 1.83e+1 (6.58e-1) & 1.75e+1 (7.03e-1) & 2.18e+1 (4.19e+0) \\
\qquad Student-$t$ & \textbf{1.77e+0 (6.65e-2)} & 3.15e+0 (1.67e-1) & 5.95e+1 (1.44e+0) & 5.20e+1 (2.19e+0) & 4.95e+1 (1.75e+0) & 6.90e+1 (2.02e+1) \\
\qquad Laplace & \textbf{1.61e+0 (4.21e-2)} & 3.51e+0 (1.93e-1) & 5.60e+1 (1.58e+0) & 4.23e+1 (1.72e+0) & 4.06e+1 (1.47e+0) & 5.62e+1 (1.66e+1) \\
\bottomrule
\end{tabular}
\end{table}

\begin{table}
    \centering
\caption{Comparison with Huber GP: Root mean squared error for 100\% Laplace noise}
\label{table:rmse_100_percent_huber}
\tiny
\begin{tabular}{lccccccc}
\toprule
Data & Standard & Relevance Pursuit & Student-$t$ & Laplace & Huber & Huber + Projection \\
\midrule
Neal \\
\qquad  Student-$t$ & 1.96e+0 (3.64e-1) & 1.56e+0 (2.48e-1) & \textbf{7.39e-1 (4.08e-2)} & \textbf{7.54e-1 (3.17e-2)} & \textbf{7.48e-1 (3.00e-2)} & \textbf{7.48e-1 (3.01e-2)} \\
\qquad Laplace & 1.51e+0 (1.06e-1) & 2.40e+0 (2.72e-1) & \textbf{1.16e+0 (9.18e-2)} & \textbf{1.12e+0 (9.90e-2)} & \textbf{1.12e+0 (9.80e-2)} & \textbf{1.12e+0 (9.80e-2)} \\
Friedman 5 \\
\qquad Student-$t$ & 1.67e+1 (1.83e+0) & 9.30e+0 (3.05e-1) & 5.26e+0 (1.30e-1) & \textbf{4.57e+0 (1.60e-1)} & \textbf{4.61e+0 (1.55e-1)} & \textbf{4.61e+0 (1.55e-1)} \\
\qquad Laplace & 1.39e+1 (5.95e-1) & 1.37e+1 (6.36e-1) & 8.33e+0 (3.69e-1) & \textbf{7.34e+0 (3.50e-1)} & \textbf{7.40e+0 (3.45e-1)} & \textbf{7.40e+0 (3.45e-1)} \\
Friedman 10 \\
\qquad Student-$t$ & 2.10e+1 (3.42e+0) & 7.80e+0 (2.03e-1) & 4.72e+0 (1.14e-1) & \textbf{4.02e+0 (9.57e-2)} & \textbf{4.05e+0 (9.34e-2)} & \textbf{4.05e+0 (9.34e-2)} \\
\qquad Laplace & 1.30e+1 (3.77e-1) & 1.27e+1 (3.99e-1) & 7.24e+0 (1.99e-1) & \textbf{6.07e+0 (1.91e-1)} & \textbf{6.26e+0 (1.71e-1)} & \textbf{6.26e+0 (1.71e-1)} \\
Yacht \\
\qquad Student-$t$ & 5.89e+1 (1.37e+1) & 2.44e+1 (1.77e+0) & \textbf{1.57e+1 (1.36e-1)} & \textbf{1.57e+1 (1.51e-1)} & \textbf{1.57e+1 (1.51e-1)} & \textbf{1.60e+1 (3.16e-1)} \\
\qquad Laplace & 2.57e+1 (1.17e+0) & 4.75e+1 (3.51e+0) & \textbf{1.64e+1 (3.18e-1)} & \textbf{1.61e+1 (2.52e-1)} & \textbf{1.62e+1 (2.60e-1)} & \textbf{1.67e+1 (3.55e-1)} \\
CA Housing \\
\qquad Student-$t$ & 2.92e+0 (5.80e-1) & \textbf{1.24e+0 (6.17e-2)} & \textbf{1.17e+0 (3.65e-3)} & \textbf{1.18e+0 (4.42e-3)} & \textbf{1.17e+0 (4.14e-3)} & 1.21e+0 (3.05e-2) \\
\qquad Laplace & 1.57e+0 (8.25e-2) & 2.06e+0 (1.44e-1) & \textbf{1.21e+0 (1.08e-2)} & \textbf{1.20e+0 (9.39e-3)} & \textbf{1.20e+0 (9.69e-3)} & 1.23e+0 (2.05e-2) \\
\bottomrule
\end{tabular}
\end{table}

\begin{table}
\caption{Comparison with Huber GP: Negative log predictive density with 100\% Laplace noise}
\label{table:nlpd_100_percent_huber}
\tiny
\begin{tabular}{lccccccc}
\toprule
Data & Standard & Relevance Pursuit & Student-$t$ & Laplace & Huber & Huber + Projection \\
\midrule
Neal \\
\qquad Student-$t$ & 1.95e+0 (1.22e-1) & 8.78e+1 (2.80e+1) & \textbf{1.17e+0 (6.08e-2)} & \textbf{1.22e+0 (5.71e-2)} & \textbf{1.20e+0 (5.17e-2)} & \textbf{1.21e+0 (5.15e-2)} \\
\qquad Laplace & 1.89e+0 (1.14e-1) & 1.38e+2 (3.98e+1) & \textbf{1.57e+0 (9.22e-2)} & \textbf{1.55e+0 (9.20e-2)} & \textbf{1.55e+0 (9.12e-2)} & \textbf{1.55e+0 (9.12e-2)} \\
Friedman 5 \\
\qquad Student-$t$ & 4.43e+0 (8.19e-2) & 3.93e+0 (2.19e-2) & 3.11e+0 (2.69e-2) & \textbf{2.96e+0 (3.49e-2)} & \textbf{2.97e+0 (3.27e-2)} & \textbf{2.97e+0 (3.27e-2)} \\
\qquad Laplace & 4.55e+0 (2.81e-2) & 4.56e+0 (1.88e-2) & 3.64e+0 (4.03e-2) & \textbf{3.46e+0 (4.45e-2)} & \textbf{3.48e+0 (4.31e-2)} & \textbf{3.48e+0 (4.31e-2)} \\
Friedman 10 \\
\qquad Student-$t$ & 4.60e+0 (8.54e-2) & 3.89e+0 (1.36e-2) & 3.04e+0 (2.45e-2) & \textbf{2.83e+0 (2.30e-2)} & \textbf{2.84e+0 (2.22e-2)} & \textbf{2.84e+0 (2.22e-2)} \\
\qquad Laplace & 4.58e+0 (1.25e-2) & 4.57e+0 (1.27e-2) & 3.60e+0 (2.53e-2) & \textbf{3.32e+0 (3.01e-2)} & \textbf{3.37e+0 (2.61e-2)} & \textbf{3.37e+0 (2.61e-2)} \\
Yacht \\
\qquad Student-$t$ & 5.14e+0 (1.85e-1) & \textbf{4.60e+0 (8.15e-2)} & 7.30e+0 (1.40e-1) & 6.79e+0 (1.52e-1) & 6.70e+0 (1.32e-1) & 1.28e+1 (7.75e-1) \\
\qquad Laplace & \textbf{4.74e+0 (4.45e-2)} & 5.30e+0 (8.31e-2) & 4.91e+0 (8.87e-2) & 5.19e+0 (1.17e-1) & 5.12e+0 (1.04e-1) & 9.90e+0 (6.44e-1) \\
CA Housing \\
\qquad Student-$t$ & 2.08e+0 (1.01e-1) & \textbf{1.79e+0 (6.40e-2)} & 7.41e+0 (2.39e-1) & 6.52e+0 (2.54e-1) & 6.41e+0 (2.07e-1) & 6.83e+0 (3.28e-1) \\
\qquad Laplace & \textbf{1.88e+0 (5.25e-2)} & 2.24e+0 (9.41e-2) & 2.92e+0 (6.01e-2) & 3.33e+0 (7.47e-2) & 3.27e+0 (5.84e-2) & 6.64e+0 (3.28e+0) \\
\bottomrule
\end{tabular}
\end{table}

\clearpage
\section*{NeurIPS Paper Checklist}

\begin{enumerate}

\item {\bf Claims}
    \item[] Question: Do the main claims made in the abstract and introduction accurately reflect the paper's contributions and scope?
    \item[] Answer: \answerYes{} 
    \item[] Justification: Everything that is claimed in the abstract and introduction is discussed in detail in the rest of the paper.

\item {\bf Limitations}
    \item[] Question: Does the paper discuss the limitations of the work performed by the authors?
    \item[] Answer: \answerYes{} 
    \item[] Justification: Limitations of the work are mentioned throughout the paper as well as summarized in the conclusion.

\item {\bf Theory Assumptions and Proofs}
    \item[] Question: For each theoretical result, does the paper provide the full set of assumptions and a complete (and correct) proof?
    \item[] Answer: \answerYes{} 
    \item[] Justification: All theoretical results clearly state the required assumptions; proofs are provided in Appendix~\ref{appdx:proofs}.

    \item {\bf Experimental Result Reproducibility}
    \item[] Question: Does the paper fully disclose all the information needed to reproduce the main experimental results of the paper to the extent that it affects the main claims and/or conclusions of the paper (regardless of whether the code and data are provided or not)?
    \item[] Answer: \answerYes{} 
    \item[] Justification: Code for our method as well as for the benchmarking setup is provided in the supplementary material, and core models and algorithms will be open-sourced upon publication.

\item {\bf Open access to data and code}
    \item[] Question: Does the paper provide open access to the data and code, with sufficient instructions to faithfully reproduce the main experimental results, as described in supplemental material?
    \item[] Answer: \answerYes{} 
    \item[] Justification: Code for reproducing the results in the paper is included in the supplementary material. All data sets are publicly available and referenced in the work and the code.

\item {\bf Experimental Setting/Details}
    \item[] Question: Does the paper specify all the training and test details (e.g., data splits, hyperparameters, how they were chosen, type of optimizer, etc.) necessary to understand the results?
    \item[] Answer: \answerYes{} 
    \item[] Justification: Details on the experiments are provided in the experimental section as well as the Appendix. Exact implementation details are contained in the included code submission. 

\item {\bf Experiment Statistical Significance}
    \item[] Question: Does the paper report error bars suitably and correctly defined or other appropriate information about the statistical significance of the experiments?
    \item[] Answer: \answerYes{} 
    \item[] Justification: Variance in the results is provided throughout in the form of confidence intervals and/or violin plots.

\item {\bf Experiments Compute Resources}
    \item[] Question: For each experiment, does the paper provide sufficient information on the computer resources (type of compute workers, memory, time of execution) needed to reproduce the experiments?
    \item[] Answer: \answerYes{} 
    \item[] Justification: Total compute resources spent are described in Section~\ref{appdx:AddEmpirical:Compute} and are estimated at around 2 CPU years in total. 
    
\item {\bf Code Of Ethics}
    \item[] Question: Does the research conducted in the paper conform, in every respect, with the NeurIPS Code of Ethics \url{https://neurips.cc/public/EthicsGuidelines}?
    \item[] Answer: \answerYes{} 
    \item[] Justification: The research fully conforms to the NeurIPS Code of ethics.

\item {\bf Broader Impacts}
    \item[] Question: Does the paper discuss both potential positive societal impacts and negative societal impacts of the work performed?
    \item[] Answer: \answerNA{} 
    \item[] Justification: The paper makes foundational methodological contributions to robust probabilistic regression, and we do not see any direct societal impacts of the work that would require specific discussion.
    
\item {\bf Safeguards}
    \item[] Question: Does the paper describe safeguards that have been put in place for responsible release of data or models that have a high risk for misuse (e.g., pretrained language models, image generators, or scraped datasets)?
    \item[] Answer: \answerNA{} 
    \item[] Justification: The paper poses no such risks.

\item {\bf Licenses for existing assets}
    \item[] Question: Are the creators or original owners of assets (e.g., code, data, models), used in the paper, properly credited and are the license and terms of use explicitly mentioned and properly respected?
    \item[] Answer: \answerYes{} 
    \item[] Justification: All creators or original owners of the code and data sets that were used in this paper are properly cited and credited. Licenses for these assets are mentioned in Sec.~\ref{appdx:AddEmpirical} and our use complies with those licenses. 

\item {\bf New Assets}
    \item[] Question: Are new assets introduced in the paper well documented and is the documentation provided alongside the assets?
    \item[] Answer: \answerYes{} 
    \item[] Justification: The only new asset introduced in this paper is the code for method and benchmarks; it does not introduce any other assets (data sets or models). The code is provided as an anonymized zip file for the reviewers. Upon publication, the code released as open source and documented according to community standards.

\item {\bf Crowdsourcing and Research with Human Subjects}
    \item[] Question: For crowdsourcing experiments and research with human subjects, does the paper include the full text of instructions given to participants and screenshots, if applicable, as well as details about compensation (if any)? 
    \item[] Answer: \answerNA{} 
    \item[] Justification: The paper does not involve crowdsourcing nor research with human subjects.

\item {\bf Institutional Review Board (IRB) Approvals or Equivalent for Research with Human Subjects}
    \item[] Question: Does the paper describe potential risks incurred by study participants, whether such risks were disclosed to the subjects, and whether Institutional Review Board (IRB) approvals (or an equivalent approval/review based on the requirements of your country or institution) were obtained?
    \item[] Answer: \answerNA{} 
    \item[] Justification: The paper does not involve crowdsourcing nor research with human subjects.

\end{enumerate}

\end{document}